\documentclass[runningheads]{llncs}
\usepackage{graphicx}
\usepackage{caption}
\usepackage{subcaption}
\usepackage{algorithmic}
\usepackage[ruled,vlined]{algorithm2e}
\usepackage{tabularx}
\usepackage{amsmath}
\usepackage{booktabs}
\usepackage[export]{adjustbox}
\usepackage{float}
\usepackage{comment}
\begin{document}
\title{
A Novel Neural Network-Based Federated Learning System for Imbalanced and Non-IID Data}
%
%
\author{Mahfuzur Rahman Chowdhury \and
\\Muhammad Ibrahim*
\orcidID{0000-0003-3284-8535}
}
\authorrunning{Chowdhury and Ibrahim}
%
\institute{Department of Computer Science and Engineering, University of Dhaka, Bangladesh 
\email{mahfuzurrahman-2020419543@cs.du.ac.bd, $^*$ibrahim313@du.ac.bd} (*Corresponding Author)\\
}

\maketitle              
\begin{abstract}
With the growth of machine learning techniques, privacy of users' data has become a major concern. Most of the machine learning algorithms rely heavily on large amount of data which may be collected from various sources. Collecting these data yet maintaining privacy policies has become one of the most challenging tasks for the researchers. To combat this issue, researchers have introduced federated learning, where a prediction model is learnt by ensuring the privacy of clients' data. 
However, the prevalent federated learning algorithms possess an accuracy and efficiency trade-off, 
especially for non-IID data. In this research, we propose a centralized, neural network-based federated learning system.   The centralized algorithm incorporates micro-level parallel processing inspired by the traditional mini-batch algorithm where the clients' devices and the server handle the forward and backward propagation respectively. We also devise a semi-centralized version of our proposed algorithm. This algorithm takes advantage of edge computing for minimizing the load from the central server, where clients handle both the forward and backward propagation while sacrificing the overall train time to some extent. We evaluate our proposed systems on five well-known benchmark datasets and achieve satisfactory performance in a reasonable time across various data distribution settings as compared to some existing benchmark algorithms. 

\end{abstract}

\section{Introduction}
\label{sec:introduction motivation}



Nowadays data are considered to be the fuel that empowers innovation and enables breakthrough achievements in machine learning, and, in particular, deep learning. However, the importance of data privacy is increasing with the demand for data. Collecting data from multiple sources can be difficult in practice and usually leads to legal and ethical issues. Nowadays, data collection has become a significant obstacle for researchers and companies to conduct research or develop machine learning products. Finding an optimal way to use data without the breaking privacy policies is thus one of most challenging tasks for machine learning researchers. 

In recent years, the number of edge devices such as smartphones, personal computers, laptops, tablets, IoT devices, and sensors in networks has increased drastically. In an idea called edge computing, end devices contribute to the training process of a machine learning model. Thus the current edge devices not only contribute in collection of data, but also in the learning process. This opens the door for federated learning \cite{cit1}, a new machine learning paradigm. Federated learning leverages the idea of building a model ensuring that data do not leave the source device, thereby giving rise to a client-server environment during the learning process. However, federated learning is criticized for its relatively poor performance when compared to traditional machine learning environment \cite{cit44}. 

\subsection{Motivation}
The artificial intelligence systems of today usually utilize data from different sources and use these data to predict the outcome without human intervention. Recommendation systems \cite{cit45}, ranking systems \cite{ibrahim2016tf}, \cite{sajid2023feature}, face recognition systems \cite{kalam2019facial}, \cite{cit48}, and spam detection systems \cite{cit49} are some examples out of many. Hence, data privacy has become a genuine concern for the mass people. Because of this, laws have been introduced by governments to prevent the disclosure of their citizens' personal data without permission. For example, the General Data Protection Regulation (GDPR) \cite{cit3} discusses the boundaries between multiple organizations to share data. 

Federated learning \cite{cit2} has been proposed to overcome the challenges associated with data privacy. However, federated learning comes with its own limitations. Current federated learning algorithms are less efficient than traditional machine learning algorithms. Multiple architectures have been proposed over the years to minimize the effectiveness and efficiency gaps between federated learning and traditional learning \cite{cit15}. 
The general idea of federated learning is that the users' data does not leave the users' devices. With the help of the idea of edge computing, the local models are trained on users' devices. After completing the local training process, the server collects all the local models and combines these into a global model. The global model thus learns from data of all the clients without needing them in a single device, thereby preserving privacy.  Federated learning has been gaining popularity in many domains, especially in healthcare sector \cite{antunes2022federated} as medical data are very sensitive and have restrictions to share.

Data across the clients may be highly imbalanced in terms of amount of the data a client possesses. Also, some class labels may occur too frequently in the dataset, while some other class labels may appear too sporadically, which is known as non-IID data scenario. Some federated learning algorithms have proven to be effective in terms of identically and independently distributed (IID) data. Unfortunately, when clients' data are distributed in a non-IID manner, most of the federated learning systems fail to yield satisfactory accuracy.

\subsection{Research Objectives}

In the current state-of-art methods of federated learning, to the best of our knowledge, there is no algorithm that works well with non-IID data in reasonable training time. The objective of this research is to overcome the current limitations regarding data distribution of federated learning systems. Specifically, this research focuses on achieving four key factors: 

\begin{itemize}

\item To develop a neural network-based federated learning system. 
\item To create a learning system that is independent of data distribution settings such as non-IID case.
\item To build a system that provides acceptable accuracy in tolerable training time. 
\item To construct a semi-centralized learning system that removes or minimizes server dependency. 
\end{itemize}

\subsection{Contributions}

This research proposes a neural network-based federated learning algorithm that incorporates a parallel training process over smaller chunks of data across multiple clients. The algorithm maintains acceptable performance while ensuring acceptable training time. The research also proposes a modified implementation of the proposed algorithm to remove the dependency from the server at the cost of training time. More specifically, below are the contributions of this research:
\begin{itemize}
\item We develop a novel parallel training method for neural network-based federated learning systems that mitigates the data distribution problem (imbalanced and non-IID data). In our algorithm, a client forward-propagates with a chunk of its data instead of its entire data and sends the loss values to the server. The server aggregates all the loss values of multiple clients, back-propagates its (global) model, and then sends the global model again to the clients. Our developed system also demonstrates satisfactory efficiency in terms of training time.
\item We develop a semi-centralized training method for our proposed algorithm 
that reduces the server dependency. The system achieves the same level of performance albeit with a longer training time.
\end{itemize}

The rest of the paper is organized as follows. Section~\ref{sec:background} discusses the background knowledge needed for this research. We also discuss the relevant research works. Section~\ref{sec:proposed}  proposes our approaches. Section~\ref{sec:results}  compares our approaches with popular state-of-art methods on five datasets using evaluation metrics like accuracy and F1-score. We also compare the training time. Here we discuss the overall findings of this research. Section~\ref{sec:conclusion} concludes the paper with the hint of future research avenues.

\section{Background Study and Literature Review}
\label{sec:background}

In this section, we present a brief overview of  machine learning, neural networks and distributed learning. We then discuss the relevant works on federated learning domain.

\subsection{Background Study}
This subsection briefly describes the terms machine learning, neural network, distributed learning, federated learning, and IID data scenario.
\subsubsection{Machine Learning}
\begin{figure}[htbp]
\centering
\includegraphics[scale=0.5]{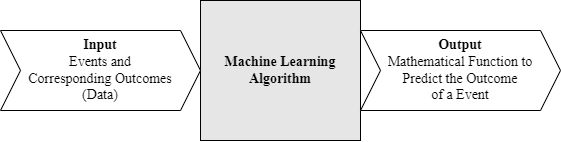}
\caption{Basic machine learning system architecture.}
\label{ML}
\end{figure}

Machine learning algorithms \cite{cit16} are effective at predicting the outcome of complex problems based on historical data. The nature of a machine learning algorithm is very different from general algorithms of computer science. Traditional algorithms take input and perform one or more mathematical operations to generate an output. Machine learning algorithms, however, learn from data to generate output. Figure \ref{ML} shows the basic working methodology of a machine learning algorithm. The algorithm takes multiple historical events and corresponding outcomes of these events as inputs, and uses these data to predict the outcomes for unseen events. Random Forest \cite{cit23}, \cite{ibrahim2022evolution}, Support Vector Machine (SVM) \cite{cit24}, Decision Tree \cite{cit25}, K-Nearest Neighborhood (K-NN) \cite{cit26} are some of the most popular machine learning algorithms.

\subsubsection{Neural Networks}
\begin{figure}[htbp]
\centering
\includegraphics[scale=0.35]{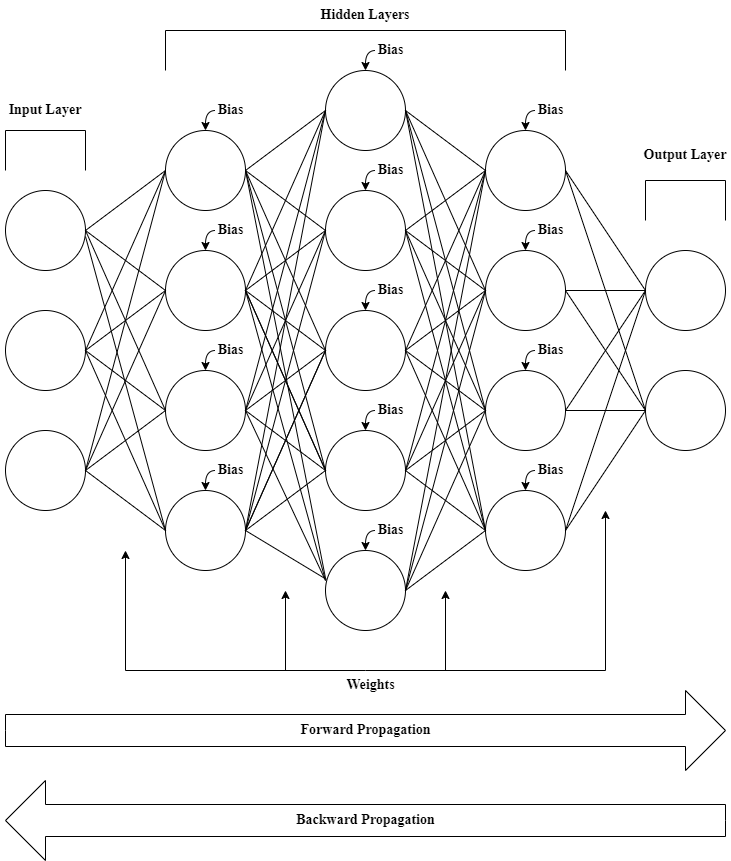}
\caption{Multi-layer neural network architecture.}
\label{NN}
\end{figure}

A neural network is a machine learning algorithm composed of multiple layers of interconnected nodes. Each of these nodes performs a simple computation. The inputs of the algorithm are fed through a layer known as input layer. The outputs are the product of the output layer. In between there are one or more hidden layers. Each of the layers performs a nonlinear operation on the output of the previous layer. During the learning process, after calculating the loss on the forward propagated data, the neurons and hidden layers update their weights and biases by using the back-propagation method. This process is repeated until the network achieves the desired level of accuracy on training data. Figure~\ref{NN} gives a example of multi-layer neural network architecture. Here the input layer has three neurons and the output layer has two neurons. There are three hidden layers consisting of four, five, and four neurons respectively. All the hidden layers' neurons have bias values in addition to weights. 

In recent years, neural networks have become tremendously popular due to the availability of increasing amount of training data and of increasing amount of computational power, thereby giving rise to the notion of deep learning. The effectiveness of neural network models is proportional to the amount of available training data~\cite{cit18} and computational resources. Neural networks are being applied to a range of applications including agriculture \cite{mimi2023identifying}, \cite{rehana2023plant}, healthcare \cite{shahid2019applications}, and economics \cite{muhammad2023transformer}.

\subsubsection{Distributed Learning}
The idea of distributed learning had been formalized and popularized in the machine learning arena in the 1990s \cite{cit19}, \cite{cit20}, \cite{cit21}.  Distributed machine learning is a method of training machine learning models where both data and computation are distributed among multiple devices or nodes instead a single device. Distributed learning enables parallel processing and minimizes the time needed to train the model. It is especially helpful when working with large datasets or computationally demanding models.

\subsubsection{Federated Learning and its Characteristics}
As mentioned in Section~\ref{sec:introduction motivation}, federated learning algorithms take the privilege of idea of edge-computing to learn a machine learning model in a client-server environment without breaching data privacy of the clients. A recent survey on federated learning systems \cite{cit2} classifies these algorithms according to five aspects. The first aspect is data partitioning, which means, how data can be distributed among different users/end devices. In horizontal distribution of data, data of all the clients or data sources have the same set of features. In vertical distribution, the clients have data on the same events but the features are different. 
There is also hybrid data distribution setting. The second aspect of this categorization is the machine learning models. Various types of machine learning models have been used in federated learning setting. The third aspect is the communication method. While the core concept of federated learning is that the data do not leave their respective devices, the clients and server communicate with each other during and after the learning process. As the main focus is privacy, this communication uses different encryption methods such as blockchain.  The fourth aspect is the communication architecture. The communication architecture is primarily of two types: one is with a server which is called centralized and the other one is without a server which is called decentralized. The fifth aspect is the scale of the federation which is primarily of two types. In cross-silo setting, the number of clients is small but the amount of data a client possesses is very large. On the other hand, in cross-device setting, the amount of data is very limited but the number of clients is large. Here the computational power of the clients is low compared to the cross-silo setting. 

\subsubsection{Imbalanced and Non-IID Data Issue}
Imbalanced data means highly disproportionate distribution of different class labels in the dataset. This characteristics of dataset poses a major problem to the accuracy of machine learning models. Therefore, practitioners need to resort to various techniques to overcome this problem \cite{ibrahim2020sampling}. IID refers to independent and identically distributed data where independent means the the data instances are not correlated with each other, and identically distributed means that the instances of all class labels are drawn from the same distribution. Assumption of this hypothesis facilitates theoretical analysis of machine learning algorithms -- such as generalization error analysis -- easier \cite{ibrahim2021understanding}. However, in practice, this assumption does not hold true in many scenarios. When it comes to the domain of federated learning, most of these algorithms perform poorly in non-IID data.

\subsection{Related Works}

We divide this section into two parts. We discuss some relevant research papers in the first part. Some federated learning algorithms that are very closely related to our research are discussed in details in the second part.

\subsubsection{Some Relevant Research Papers}

FederatedAveraging (FedAVG) \cite{cit5} is currently one of the most influential algorithms. The algorithm follows a centralized approach where the server sends the initial model to the clients, and the clients train the model with their local data. After that, the client sends the model back to the server. The server averages all the weights and bias terms to update its model. Then the server sends the model again to the clients, and this process continues for multiple iterations unless a certain level of performance is achieved by the global model. 
The authors evaluate their algorithm on two datasets, namely  MNIST and CIFAR-10. For MNIST dataset with a fully connected neural network and a convolutional neural network, accuracy of 97\% and 99\%  are reached respectively for the two architectures. 
For CIFAR-50 dataset, with a CNN model 
the performance reaches 86\% accuracy.

DfedForest algorithm \cite{cit6}   uses random forest framework \cite{ibrahim2022evolution}. The algorithm uses blockchain technology for communication between clients and the server. For this algorithm, data distribution can be horizontal or vertical. Each client builds multiple trees. After that, the client sends the tree to the server. The server sends the tree to another client for evaluation. If the accuracy of this test crosses a minimum threshold, only the tree is considered a tree of the forest, otherwise, that tree is not considered. In this way, low-performing clients can be identified. The authors use CTU-13 dataset to evaluate their proposed approach and report an F1 score of 0.98. 

Matched Average \cite{cit7} algorithm is a modified version of Fed-Avg algorithm. Here the authors use permutation invariance in each layer of each client. The authors also use the Hungarian matching algorithm \cite{cit8} to calculate the permutation and match them. Using MNIST and  CIFAR-10 datasets, the authors show that their algorithm performs better than existing popular algorithms such as Fed-AVG, FedMA, FedProx.

Fidler et al. \cite{cit9} develop a federated simulation environment for labeling medical data. In addition to this, the authors introduce a generative approach for federated learning with two neural networks.

Ziller et al. \cite{cit10} work with deep federated learning for medical image segmentation. The authors demonstrate the first application of differential private gradient descent-based federated learning on the task of semantic segmentation in computed tomography and report high segmentation performance with strong privacy guarantee and an acceptable training time.

Jiménez-Sánchez et al. \cite{cit11} investigate federated learning in the domain of breast cancer. They combine unsupervised domain adaptation to deal with domain shift while preserving data privacy. The authors evaluate their method using three clinical datasets from different vendors. Their results validate the effectiveness of federated adversarial learning for multi-site breast cancer classification.


\subsubsection{Existing Benchmark Algorithms}
This section discusses in details three benchmark algorithms of federated learning, namely, FedAVG \cite{cit5}, weighted FedAVG \cite{li2019privacy}, \cite{cit13} and cycle learning \cite{cit12}. A detailed discussion of these algorithms is necessary to understand the basic differences between our proposed methods and existing algorithms. We also highlight their strengths and limitations.

For the sake of thorough discussion, let us assume, $C = \{ c_{1}, c_{2},....,c_{n}\}$ where $C$ is the set of clients and the total number of clients is $n$, and $c_{i}$ represents $i$th client. For each client $c_{i}$, $d_{i} = \{d_{i,1}, d_{i,2},...,d_{i,m_i}\}$ where $d_{i}$ represents the data of client $c_{i}$, $d_{i,j}$ represents $j$th data instance of $C_i$, and $m_i$ represents the number of data instances $c_{i}$ has. 

\textbf{Benchmark Algorithm 1: FedAVG}
In this algorithm, a server $S$ maintains the entire learning process. Firstly, the server $S$ creates a (global) neural network architecture $M$, which means the input, output and hidden units, the initial weights and biases, their internal connections etc. After that, all the clients $C$ receive the model architecture from the server $S$ through communication channel. A client $c_{i}$ then trains the model with its own data $D_{i}$. The trained model of $C_i$ is denoted by $m_{i}$. Thus, after learning is performed in all clients, all these models share the same architecture but not the same weights and biases. After finishing the training process, the train models are sent by the clients to the server $S$. After collecting all these models, the server $S$ has a set of (local) models,  $M_{local} = \{m_{1}, m_{2},....,m_{n}\}$, which means, the server $S$ has a weight set, $W = \{W_{1},W_{2},...,W_{n}\}$ and a bias set, $B = \{B_{1},B_{2},...,B_{n}\}$ where $W_{i}$ and $B_{i}$ represents the sets of weights and biases for model $m_{i}$. In the next step, the server $S$ combines all models and creates a global model $M_{global}$. The weights and biases for $M_{global}$ are then calculated as the average of all weights and biases from sets $W$ and $B$. This weight collection and aggregation process runs multiple times to achieve a good solution. This is the working procedure of FedAvg algorithm. The pseudocode is given in Algorithm \ref{alg:FedAVG}.   

\begin{algorithm}[!ht]
\setcounter{AlgoLine}{0}
\DontPrintSemicolon 
 \textbf{Input:} Server $S$, clients set $C = \{ c_{1}, c_{2}, ...., c_{n}\}$\\
  \textbf{Output:} Global model $M_{global}$\\
  
  \KwData{Client data   $D = \{ d_{1}, d_{2}, ...., d_{n}\}$}
  $M_{global}$ $\leftarrow$ global model initialized by $S$
  
    \For{$x \leftarrow 1~ up~ to~ number \; of \; iterations$}{
        \For{$j \leftarrow 1~ up~ to~ n$  in parallel}{
            Sever sends $M_{global}$ to client $c_j$     
            
            $m_j$ $\leftarrow$ local model for client $c_j$ trained on $d_j$            
            
            Client $c_j$ sends $m_{j}$ to server $S$
        }
        $M_{global} \leftarrow \frac{1}{n} \sum_{i=1}^{n} m_i$//averaging of network weights and biases
    }
\Return $M_{global}$
\caption{FedAVG}
\label{alg:FedAVG}
\end{algorithm}

\begin{figure}[htbp]
\centering
\includegraphics[scale=0.5]{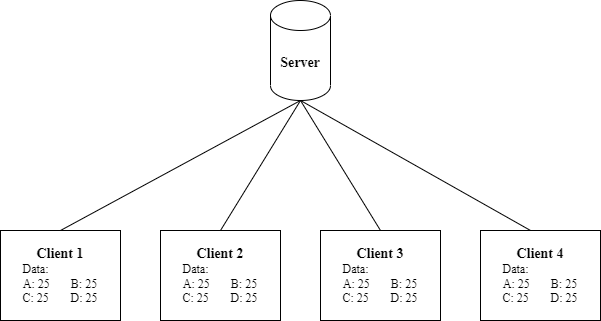}
\caption{Example of balanced and IID data}
\label{Balance and IID}
\end{figure}

FedAVG performs best under certain conditions. In particular, when the clients have an approximately equal amount of data and the data are independent and identically distributed (IID), as shown in Figure \ref{Balance and IID}, the algorithm is able to converge quickly to a high-quality global model. This observation is supported by theoretical analyses of the algorithm \cite{cit22}. 

\begin{figure}[htbp]
\centering
\includegraphics[scale=0.5]{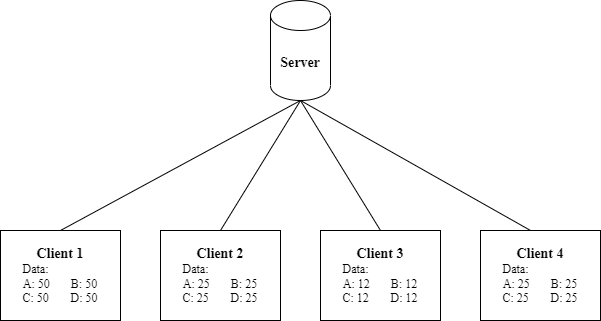}
\caption{Example of imbalanced and IID data}
\label{Imbalance and IID}
\end{figure}

Figure~\ref{Imbalance and IID} presents an example of an imbalanced  yet IID data scenario in which the participating clients have notably different amount of data. In such a scenario, FedAVG treats all clients equally, giving the same importance to each client during the model aggregation process. As a result, the updates from clients with more data, such as $M_1$, are given equal importance as the updates from clients with fewer data, such as $M_3$. For this reason, the global model $M_{global}$ is unlikely to reach a good solution. 

\textbf{Benchmark Algorithm 2: Weighted FedAVG}
Weighted FedAVG (W-FedAVG) algorithm is the result of a simple modification over FedAVG. Instead of calculating the simple average of models of all clients, this algorithm calculates a weighted average based on the amount of data a client possesses. If we compare Algorithm \ref{alg:FedAVG} (FedAVG) and Algorithm \ref{alg:WFedAVG} (W-FedAVG), we see that there are two modifications in W-FedAVG. Firstly, the algorithm takes as an additional input (through communication channel between a client and the server) the size of dataset of each client, $P = \{p_{1}, p_{2},....,p_{n}\}$. Here, $p_i$ represents the amount of data client $c_i$ has. The other modification is, in the aggregation process of Algorithm \ref{alg:WFedAVG}, W-FedAVG takes the weighted average of all local models. The weights are assigned based on the amount of local training data. 

\begin{algorithm}[!ht]
\setcounter{AlgoLine}{0}
\DontPrintSemicolon  
  \textbf{Input:} Server $S$, clients set $C = \{ c_{1}, c_{2}, ...., c_{n}\}$\\
  \textbf{Input:}Proportion of local data $P = \{p_{1}, p_{2},....,p_{n}\}$ \\
  \textbf{Output:}Global model $M_{global}$ \\
  \KwData{Client Data   $D = \{ d_{1}, d_{2}, ...., d_{n}\}$}
  $M_{global}$ $\leftarrow$ global model initialized by $S$

    \For{$x \leftarrow 1~ up~ to~ number \; of \; iterations$}{
        \For{$j \leftarrow 1~ up~ to~ n$  in parallel}{
            Sever sends $M_{global}$ to client $c_j$     
            
            $m_j$ $\leftarrow$ local model for client $c_j$ trained on $d_j$            
            
            Client $c_j$ sends $m_{j}$ to server $S$
        }
        $M_{global} \leftarrow \frac{1}{\sum_{j=1}^{n} p_j} \sum_{j=1}^{n} p_j \cdot m_j$//weighted averaging of network weights and biases
    }
    
\Return $M_{global}$
\caption{Weighted-FedAVG}
\label{alg:WFedAVG}
\end{algorithm}

\begin{figure}[htbp]
\centering
\includegraphics[scale=0.5]{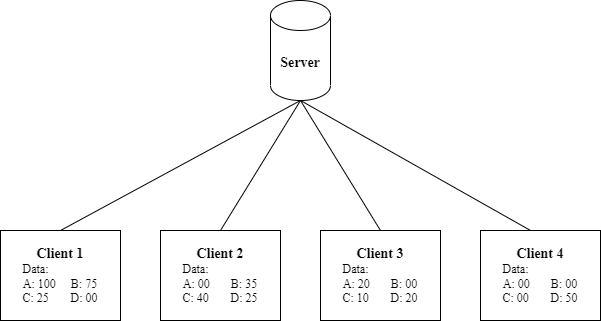}
\caption{Example of imbalanced and non-IID data}
\label{Imbalance and non-IID}
\end{figure}

Due to considering the local data sensitivity, W-FedAVG overcomes the first limitation of FedAVG which was inability to deal with imbalanced data. 
However, W-FedAVG also suffers from the issue of inability to deal with non-IID data. If we consider the scenario of Figure \ref{Imbalance and non-IID}, we see that the data across the clients are not only imbalanced but also non-IID. For example, client 1 does not have any data of class label $D$. So the local model of client 1, $m_1$ does not have any information to predict data of class label $D$. For client 2, we observe the same scenario for data of class label $A$.  So in the case of non-IID data, every local model $m_{i}$ has its limitations. Calculating an average or weighted average of their network weights and biases does not overcome this limitation. The theoretical proof of this claim is discussed in \cite{cit22}.

\textbf{Benchmark Algorithm 3: Cycle Learning}

Cycle learning (aka sequential learning) is a concept that existed even before federated learning. Cycle learning, when adapted to federated learning, overcomes the said limitations of FedAVG and weighted FedAVG. However, cycle learning is not efficient to be used in large domains due to its one-way communication and decentralized characteristics. As shown in 
Algorithm \ref{alg:CL}, in cycle learning a client passes the model to the next client after training on its data. In this way, after passing through each client, the model eventually becomes mature. However, cycle learning takes more iterations to converge to a good solution in non-IID data. The major problem of cycle learning is that the training time is huge even if the communication is flawless. If there is a large number of clients, cycle learning needs to train on all the devices one by one just to get the first version of global model, $M_{global}$. Not to mention, the whole training process is going to stall for just one communication failure between the clients. 

\begin{algorithm}[!ht]
\setcounter{AlgoLine}{0}
\DontPrintSemicolon
  \textbf{Input:} Server $S$, clients set $C = \{ c_{1}, c_{2}, ...., c_{n}\}$ \\
  \textbf{Output:} Global model $M_{global}$\\
  \KwData{Client Data   $D = \{ d_{1}, d_{2}, ...., d_{n}\}$}
  $M_{global}$ $\leftarrow$ global model initialized by $S$  

    \For{$x  \leftarrow 1 ~ up ~ to ~ number \; of \; iterations$}{
         \For{$j \leftarrow 1~ up~ to~ n$}{
            
            Client $c_j$ trains the model $M_{global}$ with its data $d_j$
            
            

            Client $c_j$ sends $M_{global}$ to client $c_{j+1}$
            
        }
    }

    
    \Return $M_{global}$
\caption{Cycle Learning}
\label{alg:CL}
\end{algorithm}

\subsection{Research Gap} 

The above discussion gives us an overview of different types of federated learning systems. From the above discussion, we identify the following research gaps:

\begin{enumerate}

\item The current state-of-the-art algorithms do not work well when the data distribution is non-IID. So there is scope for improvement in terms of accuracy for the case of non-IID data distribution. While the decentralized federated systems mitigate, albeit to some extent, this problem, these algorithms may take huge training time. 

\item Centralized federated systems (such as FedAVG and weighted FedAVG)  rely heavily on the server, whereas decentralized federated learning systems (such as cycle learning) need huge training time and suffer from communication channel vulnerability.
\end{enumerate}

In what follows, we intend to bridge these research gaps.

\section{Proposed Approach}
\label{sec:proposed}

In the previous section, we have explained a popular federated learning algorithm called FedAVG \cite{cit5} which acts as a benchmark. However, FedAvg falls short of providing an effective solution when it comes to imbalanced or non-IID data \cite{cit14}. An improved version of FedAVG algorithm, known as weighted FedAvg \cite{li2019privacy}, \cite{cit13}, partially resolves this problem with its ability to handle imbalanced data. However, in case of non-IID data, this algorithm may fail to provide a good solution \cite{cit14}. Cycle learning \cite{cit12} may be seen as a solution to this problem, but it is not a practical solution due to its high training time requirement and communication failure vulnerability \cite{cit2}.

In this section, firstly, we propose a novel neural network-based federated learning system to achieve higher accuracy for non-IID data. Secondly, we propose another version of our proposed algorithm where we minimize the server dependency (i.e., minimize the server's computational resource requirement and its network traffic) at the cost of increased training time. In particular,  we propose two federated learning systems: 1) a neural network-based centralized algorithm and 2) a semi-centralized version of our proposed algorithm. While the accuracy of these two algorithms is the same, the basic differences between them lie in training time, local computational resource utilization, and the amount of network traffic between server and clients. 

\subsection{Proposed Algorithm: A Neural Network-Based Centralized Federated Learning System}
\label{sec:proposed algo 1}
Below we discuss the motivation and strategy of our proposed algorithm, which is followed by the pseudo-code of the algorithm.

\subsubsection{Strategy}
\label{sec:our proposed algo motivation}
Here is how our proposed algorithm works. Instead of learning a local model in a client using all of its data, a client runs forward propagation on a portion of its data at one time. These clients then send their local models to the server where the backpropagation is performed after the weights and biases averaging. After the backpropagation, the server sends the global model to the clients, and the same process continues. The motivation behind this approach is that since the data can be imbalanced and non-IID, it may be better not to learn the local models using a clients all data, rather chunks of data can be used to train so that under-represented clients and class labels get a ``chance to speak'' in the global model. 

\begin{figure}[htbp]
\centering
\includegraphics[scale=0.5]{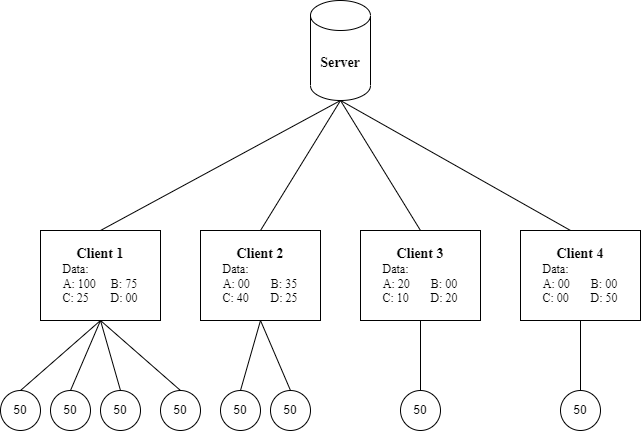}
\caption{Example of client data partitioning in our proposed method.}
\label{PM}
\end{figure}

As an example of how our proposed idea works, we consider the situation depicted in Figure \ref{Imbalance and non-IID}. Here we have four clients and  four class labels (A, B, C, and D) in the data. However, all clients do not have similar amount of data instances, nor do they have similar distribution of instances of all class labels -- implying that the data scenario here is imbalanced and non-IID. Now we explain the details steps of our algorithm. At first, all the clients send to the server the number of data instances they have. In this case, the numbers are 200, 100, 50, and 50. Here we assume the values of two hyper-parameters, the batch size and the parallel training window, are 100 and 2, respectively. Now the server sends an instruction to the clients, which is, to divide their data into groups of $100 / 2 = 50$. This scenario is depicted in Figure \ref{PM}. After dividing their data accordingly, all clients may not have the same number of groups of data. Also, the information as to which clients would take part in parallel in the training should be communicated by the server. To facilitate this communication, the server's instruction also contains information as to which clients need to take part in training in which iteration. Then, according to this information at hand, the server initiates the training process by sending the global neural network model architecture to the designated clients. As the parallel training window size is 2, at a time maximum two clients train (forward-propagate) the model with a chunk of their data, and, afterwards, send their loss values to server. After getting these loss values from clients, the server calculates the average loss, updates (back-propagates) the global model, and then repeats the process.

\begin{figure}[htbp]
\centering
\includegraphics[scale=0.25]{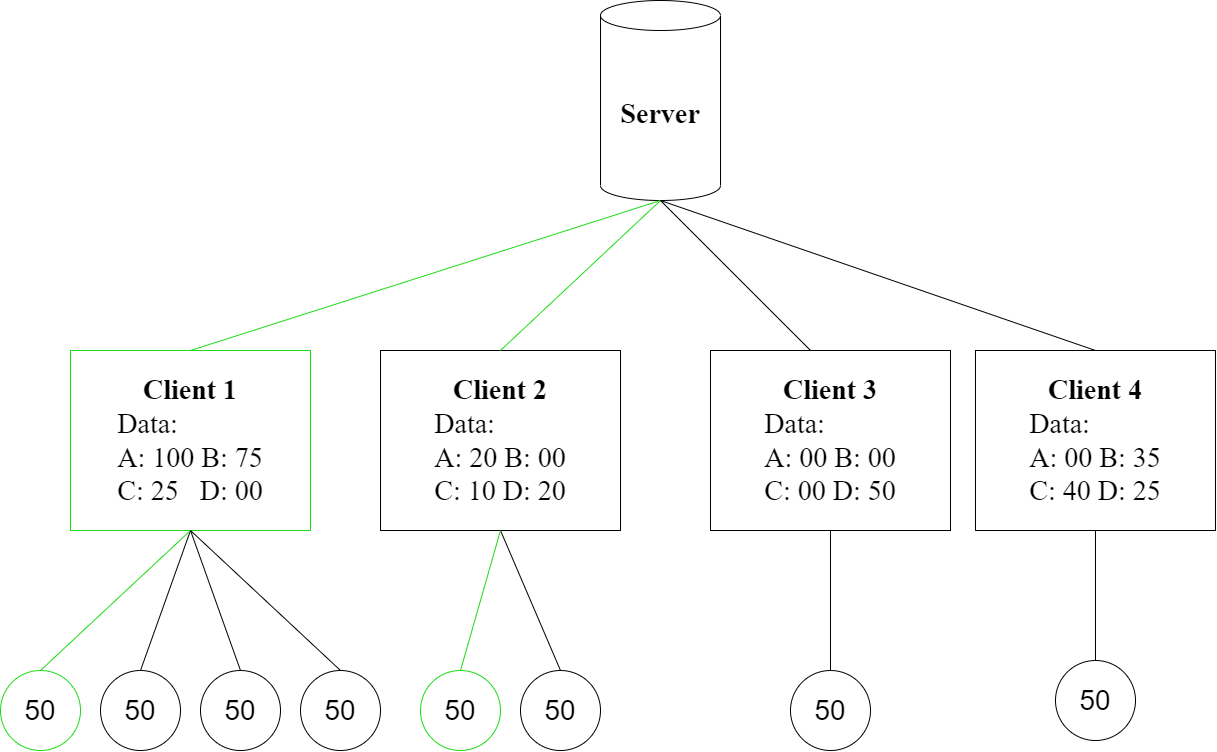}
\caption{Example of client data training flow in our proposed method: Step 1}
\label{PTrain1}
\end{figure}

\begin{figure}[htbp]
\centering
\includegraphics[scale=0.25]{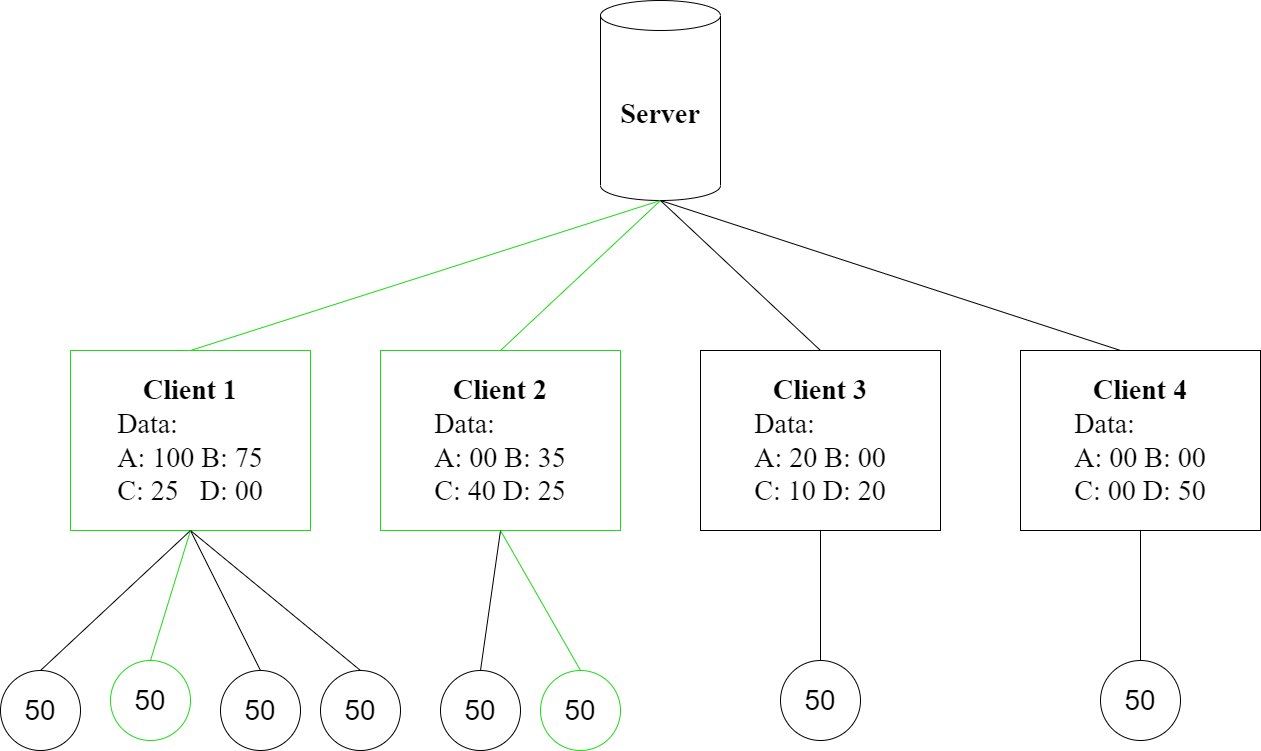}
\caption{Example of client data training flow in our proposed method: Step 2}
\label{PTrain2}
\end{figure}

\begin{figure}[htbp]
\centering
\includegraphics[scale=0.25]{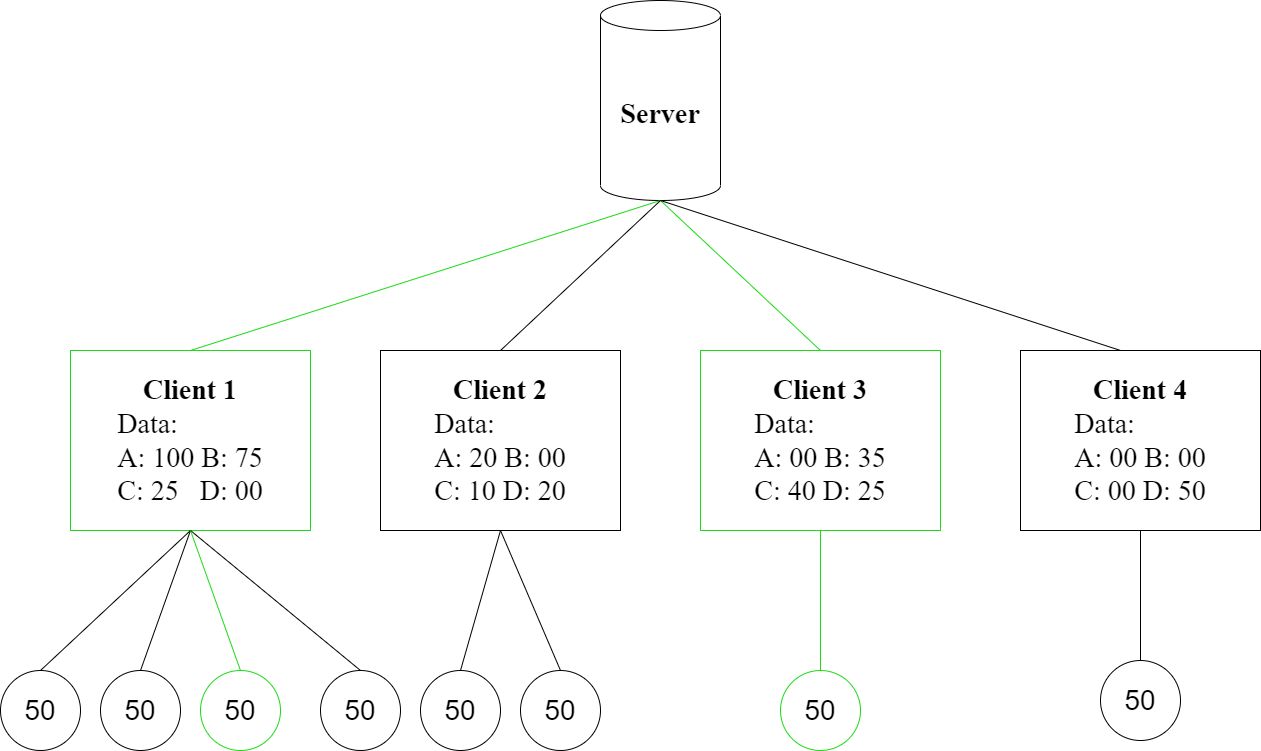}
\caption{Example of client data training flow in our proposed method: Step 3}
\label{PTrain3}
\end{figure}

\begin{figure}[htbp]
\centering
\includegraphics[scale=0.25]{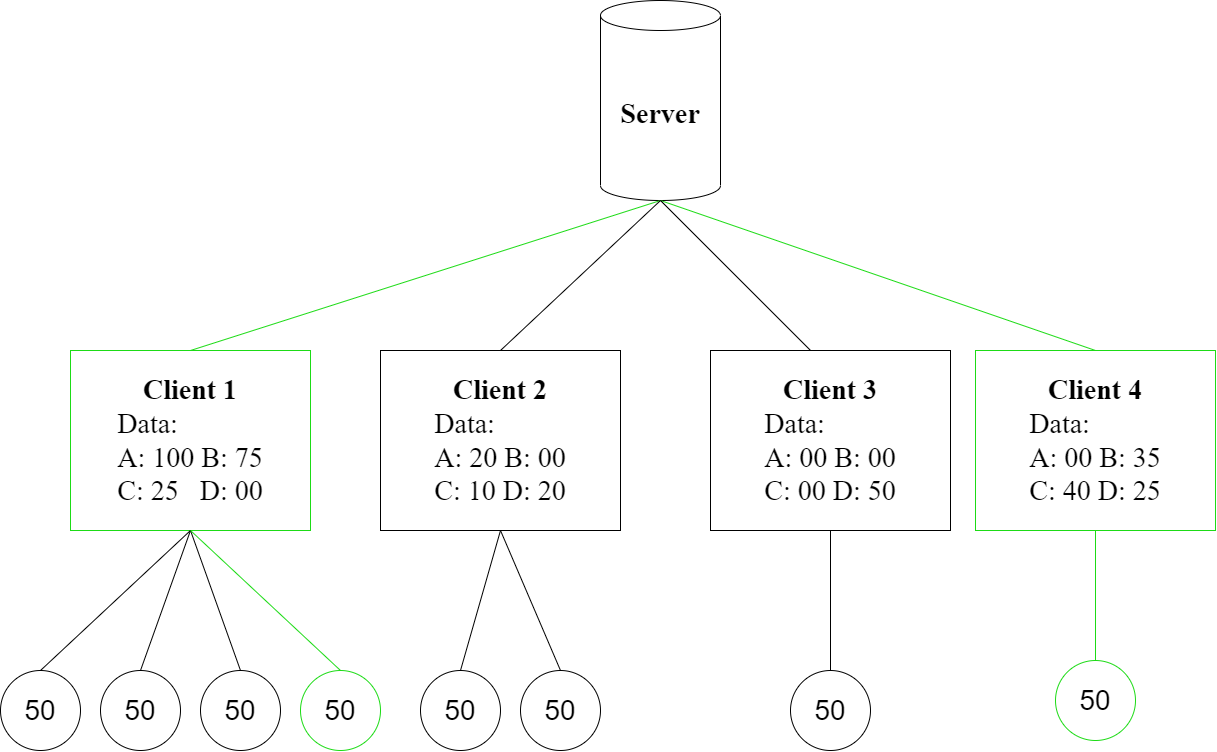}
\caption{Example of client data training flow in our proposed method: step 4}
\label{PTrain4}
\end{figure}

To continue with our running example, we now consider Figures \ref{PTrain1}, \ref{PTrain2}, \ref{PTrain3}, and \ref{PTrain4}. In Figure \ref{PTrain1}, we see the first step: the server communicates with only clients 1 and 2 because the parallel training window size is 2. Client 1 trains the model with only one of its 4 groups of data because the batch size is 50. The same goes for client 2. After finishing the forward propagation, both of these clients send their loss values to the server along with the parameters of the network. The server calculates the average loss, updates the global model accordingly. In Figure \ref{PTrain2}, the server sends the updated model again to clients 1 and 2. Now the clients train their models with another group of data. This process continues in Figures \ref{PTrain3} and \ref{PTrain4}. This way a single iteration of the algorithm is executed. After completing the entire training process with multiple iterations, the global model $M_{global}$ is fully learnt.

There are two hyper-parameters in our algorithm: the batch size and the parallel training window size. The batch size indicates the amount of data that would be feed-forward before one back-propagation. The parallel training window size is the number of clients that would feed-forward the model before a back-propagation. Here is how these two hyper-parameters are set. In the first phase, the server collects the data count from every client. The server then sorts the clients according to their data count. In the next step, the server calculates the amount of data that the a client needs to feed-forward before sending its loss values to the server. Here the server selects the batch size divided by the parallel training window size as the amount of data a client should feed-forward its local model.

By now we know that our proposed strategy to solve the problem can be divided into two parts: the instruction generation process and the training process. In the instruction generation phase, server generates instructions for every client where the information is mentioned as to which clients would take part in the training and with how many data instances. Every client follows the instructions in the next phase which is the actual training (forward-propagation) process.

Now we address a vital question: why our proposed method is supposed to be effective (1) in imbalanced data scenario, and (2) in non-IID data scenario? To answer this question, let us consider the following arguments. The method should work well for imbalanced data scenario because in the early phase of learning we assign equal importance to all the clients as long as we can, i.e., as long as a client is not exhausted with its (local) data. However, as the data of some clients are exhausted, the heavy clients gradually get more importance in the learning process because they still have data at their hands. This is in contrast with FedAVG and other centralized non-weighted methods because they give equal importance to all clients all the time of the entire learning process. Now to address the second part of the question, we argue that our method should work well for non-IID data scenario. The arguments to support this conjecture are as follows. In non-IID data, the clients have highly skewed class distribution in their (local) datasets. Because of this property, a local model trained on non-IID data can be heavily biased towards certain classes. For example, in Figure \ref{Imbalance and non-IID}, for client 1, 50\% of the data belong to one class which is class A, but class C has a very small amount of data, and class D has no data at all. So the weights and biases of the local models across multiple clients are very different from each other. If we want to remove this problem in traditional learning, what we do is that we collect all the data and shuffle them before training. In the traditional  learning process of neural networks, people use the mini-batch method. This means that after feed-forwarding a certain amount of data (also referred to as a batch), the loss is calculated and then back-propagated to update the weights and bias terms. Because of shuffling the whole data before forward-propagation, each batch contains similar distribution of all class labels. So when the loss is calculated after the forward pass with a batch of data, the loss is representative of all class labels. In our algorithm, we take inspiration from this shuffling idea. To adapt this idea to the non-IID data scenario of federated learning, we forward-propagate with a chunk of data of some clients instead of using all of their data.

\subsubsection{Algorithm}
We now present the algorithmic aspects of our proposed method. As explained above, the steps of our proposed algorithm can be divided into two parts: instruction generation phase and the training phase. 

\textbf{Phase 1: Instruction Generation}

Recall that the two hyper-parameters are: 1) the $batch\_size$ which is the amount of data a client trains on its local model, and 2) the $parallel\_window\_size$ which is how many clients train their local models at a time. Algorithm \ref{alg:IG} details the process of generating instructions by the server for the clients.

\begin{algorithm}[!ht]
\setcounter{AlgoLine}{0}
\DontPrintSemicolon  
\textbf{Input:} Server $S$, clients set $C = \{ c_{1}, c_{2},....,c_{n}\}$, client data $D = \{ d_{1}, d_{2},....,d_{n}\}$ \\
\textbf{Input:} batch\_size, parallel\_window\_size\\
\textbf{Output:} Instructions set $Ins$\\
 
    $S$ collects the sizes of datasets residing at all clients $C$

  $D_{sorted} \leftarrow  Descending\_Sort(D) $

  $Ins \leftarrow  empty\; list$  
   
  \For{$d \leftarrow \; D_{sorted}$}{
  
        $first\_train \leftarrow first\_train + (d - (d \mod \frac{batch\_size}{parallel\_window \_size} ))$  
        
  }

   $Collection \leftarrow [0] \times \frac{first\_train}{batch\_size}$
  
  \For{$d \leftarrow \; D_{sorted}$}{

        $d_{ins} \leftarrow  empty\; list $

        \If{$d \geq \frac{batch\_size}{parallel\_window \_size}$}
    {
        $d_{ins}.append(\frac{batch\_size}{parallel\_window \_size})$
    }
    
     \Else
    {
    	$d_{ins}.append(d)$
    }
    
    \If{$d \mod \frac{batch\_size}{parallel\_window \_size} = 0$}
    {
        $d_{ins}.append(0)$
    }
    
     \Else
    {
    	$d_{ins}.append(1)$
    }
    
    $d_{new} \leftarrow  d$

    $count \leftarrow  0$

     \While{$count \leq len(Collection)$}
    {

        \If{$collection[count] < batch\_size\ and\ d_{new} \ge \frac{batch\_size}{parallel\_window \_size} $}
         {       
            $d_{ins}.append(1)$
            
            $d_{new} \leftarrow d_{new} - \frac{batch\_size}{parallel\_window \_size} $
            
            $collection[count] \leftarrow collection[count] +  \frac{batch\_size}{parallel\_window \_size} $
        }
        
        \Else{
            $d_{ins}.append(0)$
            }

        $count \leftarrow count + 1$

    }
        
    $Ins.append(d_{ins})$

  }    

\Return {$Ins$}
\caption{Proposed Algorithm's Phase 1: Instruction Generator}
\label{alg:IG}
\end{algorithm}

Now let us elaborate the steps of Algorithm~\ref{alg:IG}. In Line 1, the server $S$ collects the sizes of datasets residing at  all the clients.   
 Lines 2 and 3 create two separate lists where the first one is a sorted list based on the amount of data each client has, and the other is an empty list. Lines 4 and 5 execute a for loop where the training size before each back-propagation is determined by the equation in line 5. Lines from 7 up to 27 are responsible for instruction generation. Lines 8 to 16 check two major factors: a) whether or not the client has enough remaining data for contributing to the training, and d) whether or not the iteration needs more clients to participate. After calculating these factors, the algorithm moves to lines 19 up to 26 which executes a for loop. This for loop creates the main instruction array for each client using the equations in lines 22 and 23.

\textbf{Phase 2: Training Process}

Here the server starts training the model. After generating all the instructions for the training process as per Algorithm~\ref{alg:IG}, the server first instructs the clients to divide their data into groups specified in the instruction set. After that, the server reads the individual instructions for clients and accordingly sends the global model to the nominated clients. If a client receives the (global) model from the server, the client feed-forwards one of its unused groups of data through the model. After that, the client sends the loss values generated by feed-forwarding its group of data. In Algorithm \ref{alg:TP}, lines 6 to 10 describe the process. After receiving all the loss values from the nominated clients, the server aggregates the loss values, updates (i.e., back-propagates) the weights and biases of the model, and then repeats the process. After completing all the instructions, i.e., after exhausting all the data of all the clients, a single iteration of the training is performed. This way a predefined number of training iterations is performed, thereby resulting in the final global model $M_{global}$. 

\begin{algorithm}[!ht]
\setcounter{AlgoLine}{0}
\DontPrintSemicolon
  
\textbf{Input:} Server $S$, clients set $C = \{ c_{1}, c_{2},....,c_{n}\}$\\
\textbf{Input:} Client data $D = \{ d_{1}, d_{2},....,d_{n}\}$\\
\textbf{Input:} batch\_size, parallel\_window\_size\\
\textbf{Output:} Global Model $M_{global}$\\

  $M \leftarrow Initial\ Model$

  $Instruction \leftarrow Instruction\_Generator(S, C, D, batch\_size, parallel\_window \_size)$

  $times \leftarrow 0$
  
  \For{$i \leftarrow len(Instruction)$}{

    $ins \leftarrow Instruction[i]$
    
    \If{$ins[times] = 1$}{
    
        $S\ Sends\ M\ to\ c_{i}$

        $c_{i}\ train\ M\ with\ one\ of\ its\ unused\ data\ group\ from\ d_i$

        $l_i \leftarrow calculating\ the\ loss\ of \; M$

        $c_{i}\ sends\ l_i\ to\ S \; with \; parameters$
        
    }
    \Else{

        $l_i \leftarrow 0$
        }

    $S\ receives\ all\ loss\ set\ l$

    $t_{loss} \leftarrow {\sum_{i=1}^{n} l_i}$

    $S\ updates\ M\ by\ backpropagation\ using\ t_{loss}$

    $times \leftarrow times+1$    
    
  }
  $M_{global} \leftarrow M$
    
\Return $M_{global}$
\caption{Proposed Algorithm's Phase 2: Training Process}
\label{alg:TP}
\end{algorithm}

\subsection{A Variation of Our Proposed Algorithm}
\label{sec:proposed algo 2}
We now propose a variation of our algorithm with an aim to reduce server dependency so as to make the algorithm less centralized. In what follows, we modify Algorithms \ref{alg:IG} and \ref{alg:TP} to achieve this goal.

\subsubsection{Strategy}

One of the drawbacks of our proposed algorithm is that it is a fully centralized system, i.e., the system is heavily dependent on the server. While this approach gives us the advantage of parallellization of the training process across multiple clients thereby  reducing the training time, the server needs to employ a lot of computational resources for collecting the loss values from clients, updating the global model by aggregating them, and sending the global model back to the clients.  However, in some cases, servers may not possess such computational resources. Also, a minor technical problem in the server causes termination of the entire training process. Because of this reason, we now propose a variation of our proposed algorithm that alleviates some of the burden of back-propagation from the server and, instead, distributes this task to some selected clients. That being said, this approach mitigates the advantage of parallel training which may significantly increase the overall training time.

\begin{figure}[htbp]
\centering
\includegraphics[scale=0.25]{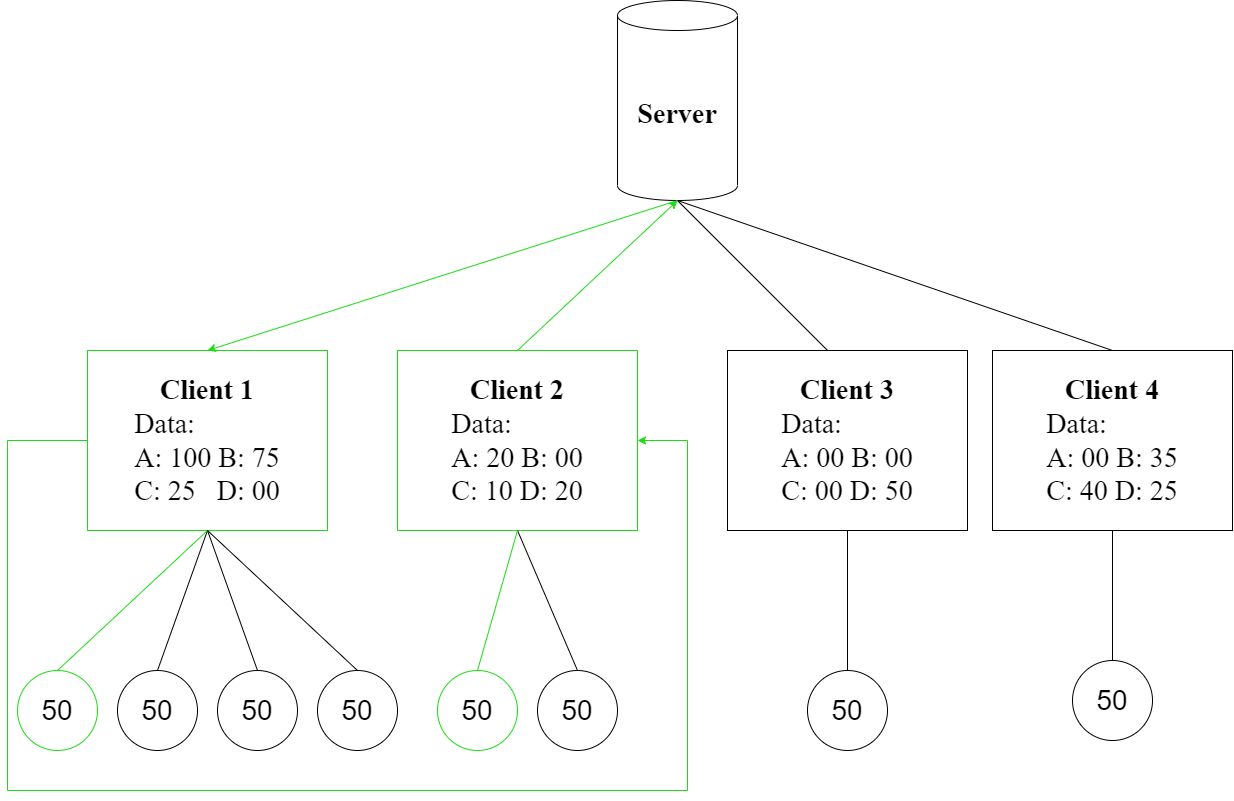}
\caption{Example of client data training flow in our proposed \textbf{variation} method: Step 1}
\label{PTrain1n}
\end{figure}

\begin{figure}[htbp]
\centering
\includegraphics[scale=0.25]{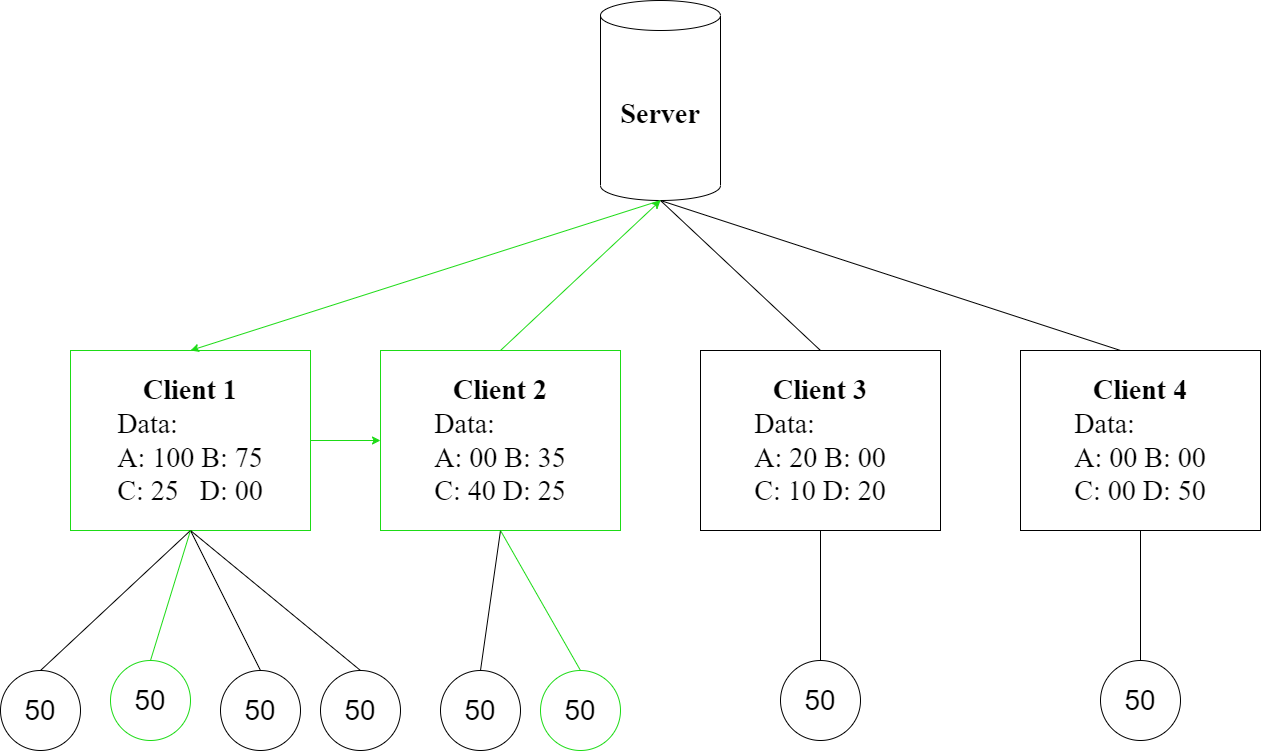}
\caption{Example of client data training flow in our proposed \textbf{variation} method: Step 2}
\label{PTrain2n}
\end{figure}

\begin{figure}[htbp]
\centering
\includegraphics[scale=0.25]{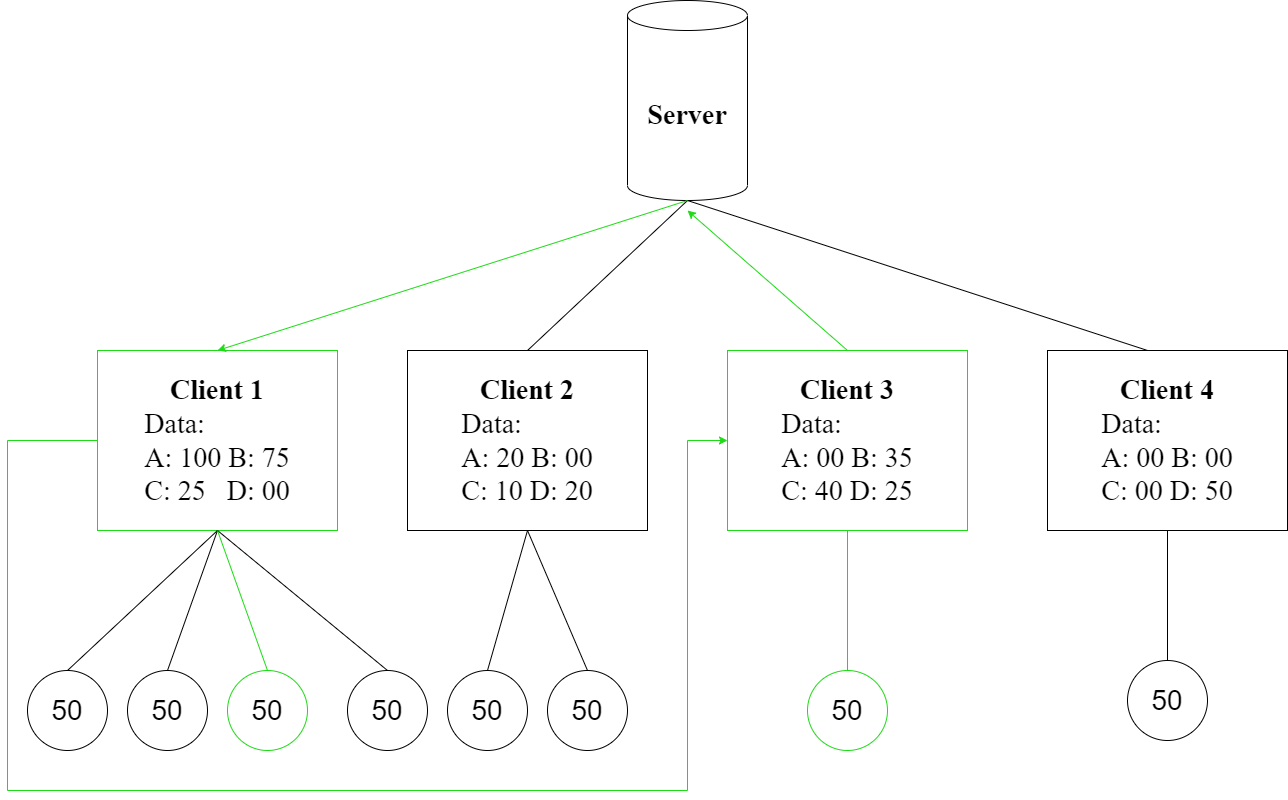}
\caption{Example of client data training flow in our proposed \textbf{variation} method: Step 3}
\label{PTrain3n}
\end{figure}

\begin{figure}[htbp]
\centering
\includegraphics[scale=0.25]{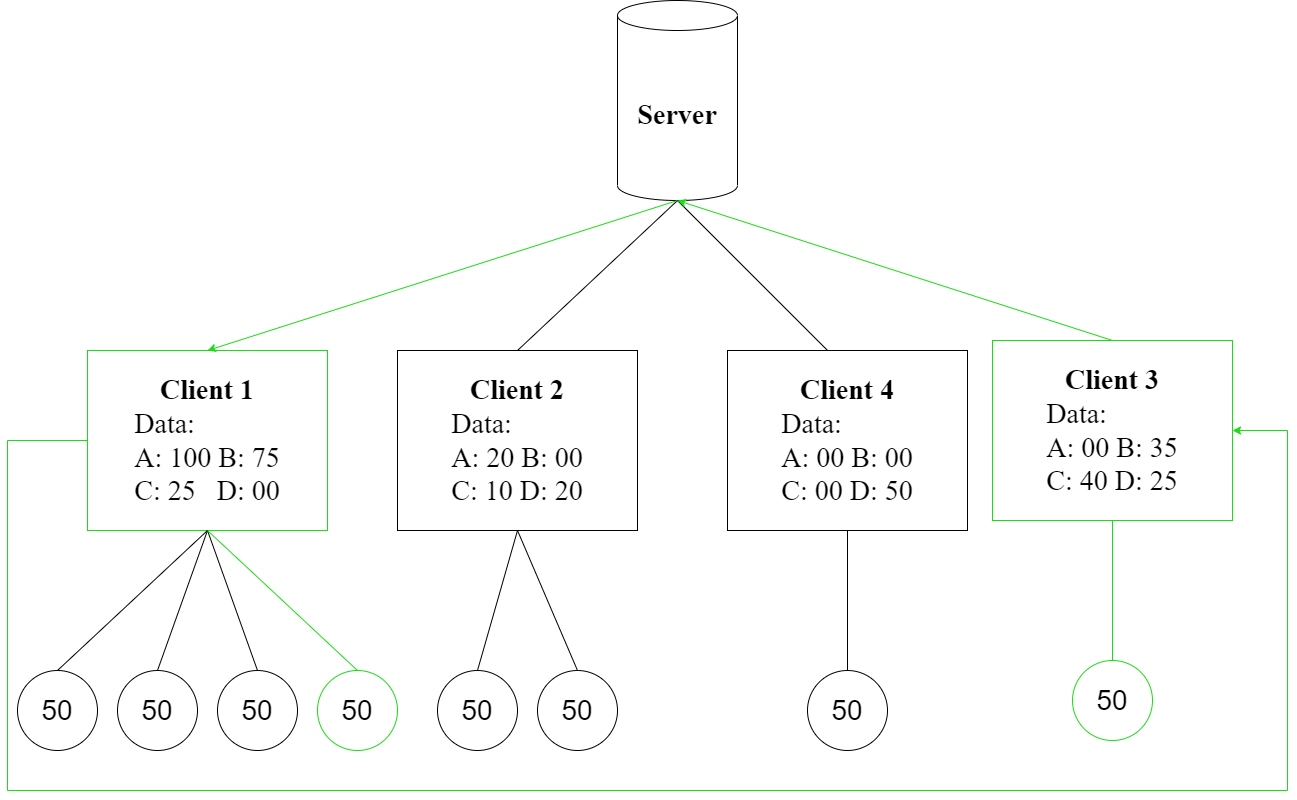}
\caption{Example of client data training flow in our proposed \textbf{variation} method: Step 4}
\label{PTrain4n}
\end{figure}

To explain our idea, let us consider Figures \ref{PTrain1n}, \ref{PTrain2n} \ref{PTrain3n} and \ref{PTrain4n}. In Figure \ref{PTrain1n}, we see the first step of our methods. The server is communicating with only client 1. Client 1 is communicating with client 2, and finally, client 2 is communicating with the server. The server first sends the model to client 1. Client 1 feed-forwards the model with its chunk of  data. After that,  client 1 sends the with loss and parameters to client 2. After that client 2 feed-forwards the original model with its own data and calculates the loss. Then client 2 aggregates the loss values and performs the back-propagation. In this way, client 2 creates the final model of step 1 and sends back the model to server. In Figure \ref{PTrain2n}, the server sends the updated model again to client 1 and the process is repeated. But it is noteworthy that now both of these clients train the model with another group of data. This processing is continued in Figures \ref{PTrain3n} and \ref{PTrain4n}. This way, after completing the full training process the global model $M_{global}$ is created.

\subsubsection{Algorithm}
We now explain the algorithmic aspects of our variation algorithm. As before, the method can be divided into two parts: instruction generation and training process.

\textbf{Phase 1: Instruction Generation and Host Client Identification}

The server collects the data size information from all the clients. After that, the server sets the $batch size$ which is how much data is going to be passed through the network before the weights and bias update. The server also determines the $cluster\_window\_size$, which is, how many clients would train the model before an update of the global model $M_{global}$. 
The Algorithm \ref{alg:IG2} is a slightly modified version of Algorithm \ref{alg:IG}. All the lines are the same with two exceptions. We use  $cluster\_window\_size$ instead of $parallel\_window\_size$ as there is no parallel training in this algorithm. Secondly, lines 23 and 24 determine the host clients which means that the client is responsible for back-propagation and for sending the model back to the server. 

\begin{algorithm}[!ht]
\setcounter{AlgoLine}{0}
\DontPrintSemicolon
  
\textbf{Input: } Server $S$, clients set $C = \{ c_{1}, c_{2},....,c_{n}\}$, client data $D = \{ d_{1}, d_{2},....,d_{n}\}$\\
  \textbf{Input: } batch\_size, cluster\_window\_size\\
  \textbf{Output: } Instructions set $Ins$, Host client set $hc$\\

  $D_{sorted} \leftarrow  Descending\_Sort(D) $

  $Ins \leftarrow  empty\; list$  
   
  \For{$d \leftarrow \; D_{sorted}$}{
        $first\_train \leftarrow first\_train + (d - (d \mod \frac{batch\_size}{cluster\_window \_size} ))$       
  
  }

   $Collection \leftarrow [0] \times \frac{first\_train}{batch\_size}$
  
  \For{$d \leftarrow \; D_{sorted}$}{

        $d_{ins} \leftarrow  empty\; list $

        \If{$d \geq \frac{batch\_size}{cluster\_window \_size}$}
    {
        $d_{ins}.append(\frac{batch\_size}{cluster\_window \_size})$
    }
    
     \Else
    {
    	$d_{ins}.append(d)$
    }
    
    \If{$d \mod \frac{batch\_size}{cluster\_window \_size} = 0$}
    {
        $d_{ins}.append(0)$
    }
    
     \Else
    {
    	$d_{ins}.append(1)$
    }
    
    $d_{new} \leftarrow  d$

    $count \leftarrow  0$
    
    \While{$count \leq len(Collection)$}
    {

        \If{$collection[count] < batch\_size\ and\ d_{new} \ge \frac{batch\_size}{cluster\_window \_size} $}
         {       
            $d_{ins}.append(1)$

            $d_{new} \leftarrow d_{new} - \frac{batch\_size}{cluster\_window \_size} $
            
            $collection[count] \leftarrow collection[count] +  \frac{batch\_size}{cluster\_window \_size} $

            \If{$collection[count] < batch\_size$}{
                $hc.append(client[d])$
            }
        }
        
        \Else{
            $d_{ins}.append(0)$
            }        

        $count \leftarrow count + 1$

    }
        
    $Ins.append(d_{ins})$
    
  }  
  
\Return $Ins$ and $hc$

\caption{Proposed Algorithm's Variation's Phase 1: Instruction Generation and Host Client Identification}
\label{alg:IG2}
\end{algorithm}


\textbf{Phase 2: Training Process}

The server initiates training the model. After generating all the instructions and determining the host clients for the training process, the server first instructs the clients to divide their data into groups. 
If a client receives an instruction from the server that means the client is going to take part in that iteration. The first client from the instruction set starts the training. After finishing the training, the client sends the model to the next client. In this order, clients complete the forward propagation part. After that, the host client is the last client to do the back-propagation. After completing the back-propagation, the client sends the model back to the server. In Algorithm \ref{alg:TP2}, lines from 6 up to 10 describe the process. 

\begin{algorithm}[!ht]
\setcounter{AlgoLine}{0}
\DontPrintSemicolon
  
  \textbf{Input: } Server $S$, clients set $C = \{ c_{1}, c_{2},....,c_{n}\}$\\
  \textbf{Input: } Clients data $D = \{ d_{1}, d_{2},....,d_{n}\}$\\
  \textbf{Input: } Host clients, $hc = \{hc_{1}, hc_{2},....,hc_{ins}\}$\\
  \textbf{Input: } batch\_size, cluster\_window\_size\\
  \textbf{Output: } Global Model $M_{global}$\\

  $M \leftarrow Initial\ Model$

  $Instruction \leftarrow Instruction\_Generator(S, C, D, batch\_size, cluster\_window \_size)$

  $times \leftarrow 0$
  
  $t_{loss} \leftarrow 0$ 
  
  $p_{parameters} \leftarrow \{\}$
  
  \For{$i \leftarrow len(Instruction)$}{

    $ins \leftarrow Instruction[i]$
    
    \If{$ins[times] = 1$}{
    
        $S\ Sends\ M\, t_{loss}\, p_{parameters}\ to\ c_{i}$

        $c_{i}\ train\ M\ with\ one\ of\ its\ unused\ data\ group\ from\ d_i$
        
        $p_{parameters}.append(save\ training\ information)$
        
        $l_i \leftarrow calculating\ the\ loss\ of \; M$

        $t_{loss} \leftarrow t_{loss} + l_i$

        \If{$c_{i} = hc_{i}$}{
            $c_{i}\ updates\ M\ by\ backpropagation\ using\ t_{loss}$
        }

        $c_{i}\ sends\ l_i\ to\ S \; with \; parameters$
        
    }    

    $times \leftarrow times+1$    
    
  }
  $M_{global} \leftarrow M$
    
\Return $M_{global}$
\caption{Proposed Algorithm's Variation's Phase 2: Training Phase}
\label{alg:TP2}
\end{algorithm}

In this section, we have proposed a neural network-based, centralized federated learning system where micro-level parallel training is introduced to improve accuracy in the case of imbalanced and non-IID data scenario. We also design a semi-centralized version of our proposed system  to reduce server dependency by sacrificing training time. In the next section, we are going to evaluate our proposed methods on multiple datasets and compare the results with existing benchmark algorithms.

\section{Performance Evaluation}
\label{sec:results}
This section presents the experimental results of our proposed two methods evaluated on five datasets. We compare our results with three benchmark algorithms, namely FedAVG, weighted FedAVG (W-FedAVG), and cycle learning. We provide a comprehensive analysis of the results to evaluate the efficiency and effectiveness of our proposed methods.

\subsection{Datasets}
We select five publicly available datasets for our evaluation, namely MNIST \cite{cit29}, Fashion MNIST \cite{cit30}, MangoLeafBD \cite{cit28}, HAM10000 \cite{tschandl2018ham10000}, and CIFAR10 \cite{krizhevsky2009learning}. 

\begin{figure}[htbp]
\centering
\includegraphics[scale=0.5]{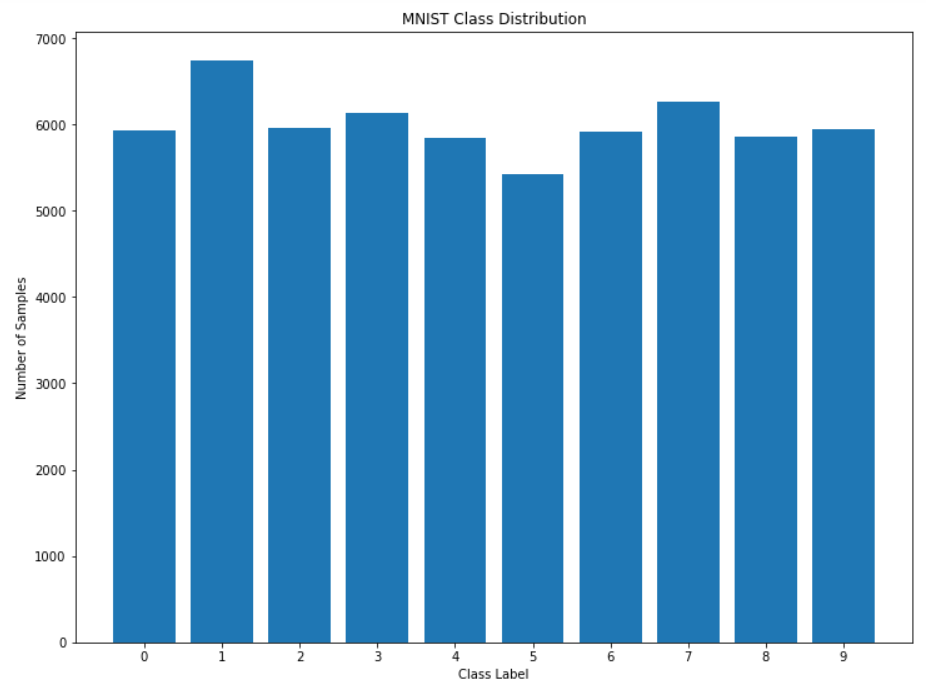}
\caption{Data distribution of MNIST dataset}
\label{MNIST Distribution}
\end{figure}

\begin{figure}[htbp]
\centering
\includegraphics[scale=0.35]{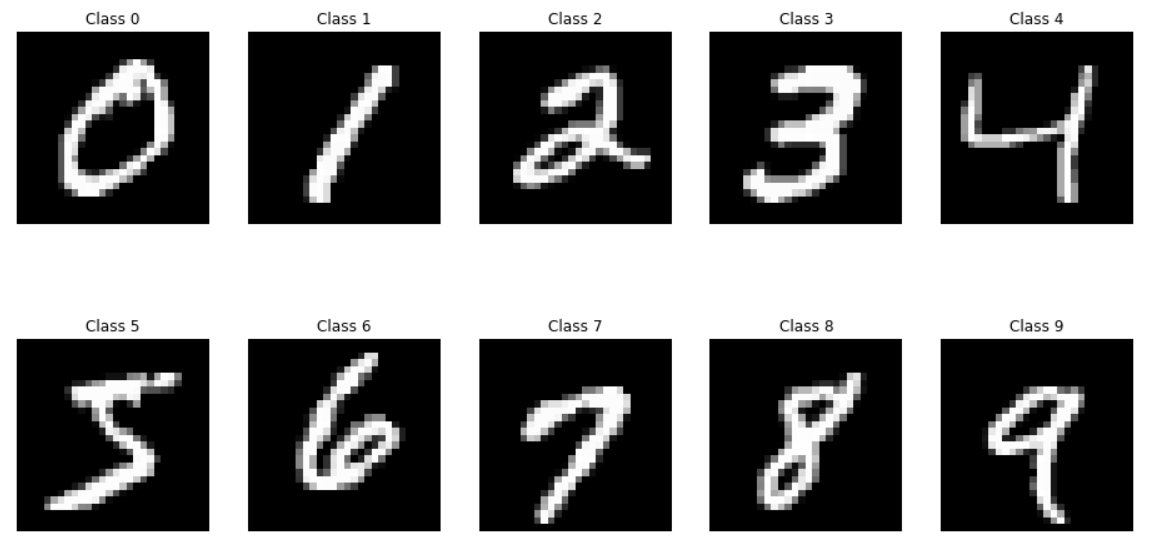}
\caption{Sample images from MNIST dataset}
\label{Sample MNIST}
\end{figure}

\begin{figure}[htbp]
\centering
\includegraphics[scale=0.35]{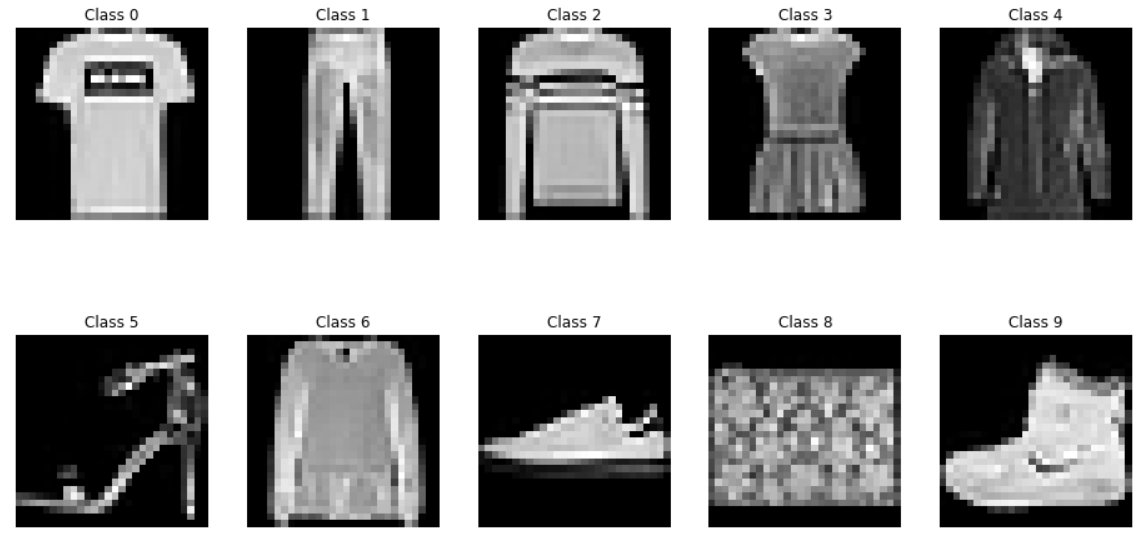}
\caption{Sample images from Fashion MNIST dataset}
\label{Sample FMNIST}
\end{figure}

\begin{figure}[htbp]
\centering
\includegraphics[scale=0.35]{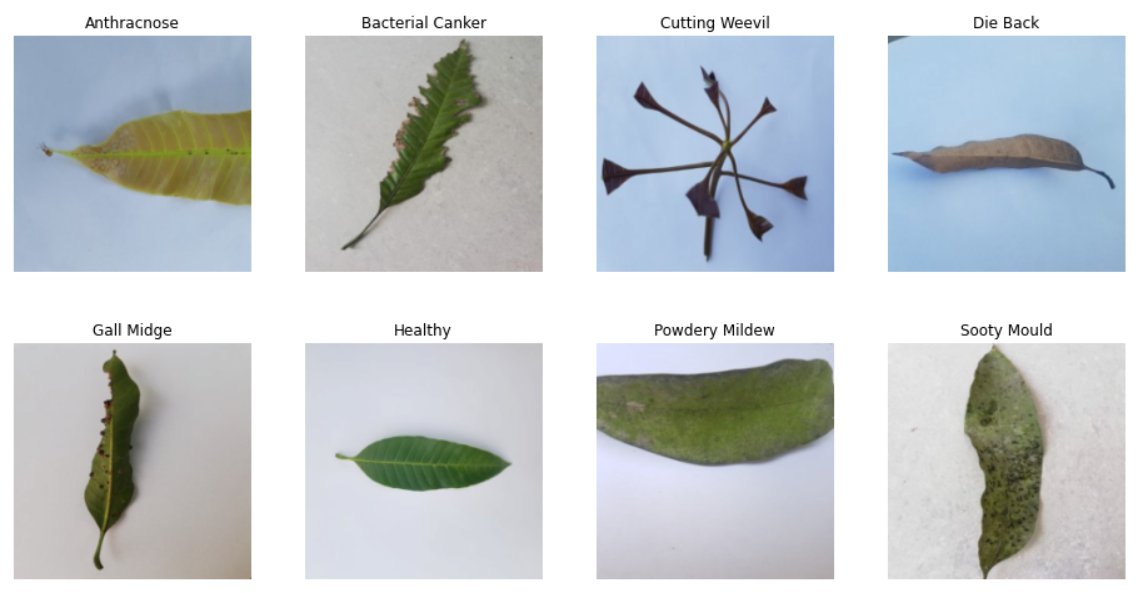}
\caption{Sample images from MangoLeadBD dataset}
\label{Sample MangoLeadBD}
\end{figure}

\begin{figure}[htbp]
\centering
\includegraphics[scale=0.5]{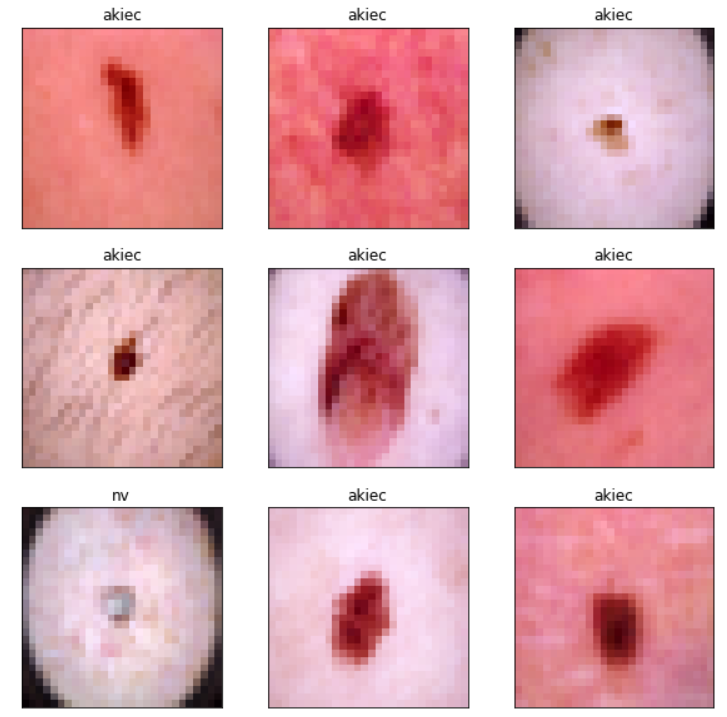}
\caption{Sample images from HAM10000 dataset}
\label{HAM7}
\end{figure}

\begin{figure}[htbp]
\centering
\includegraphics[scale=0.35]{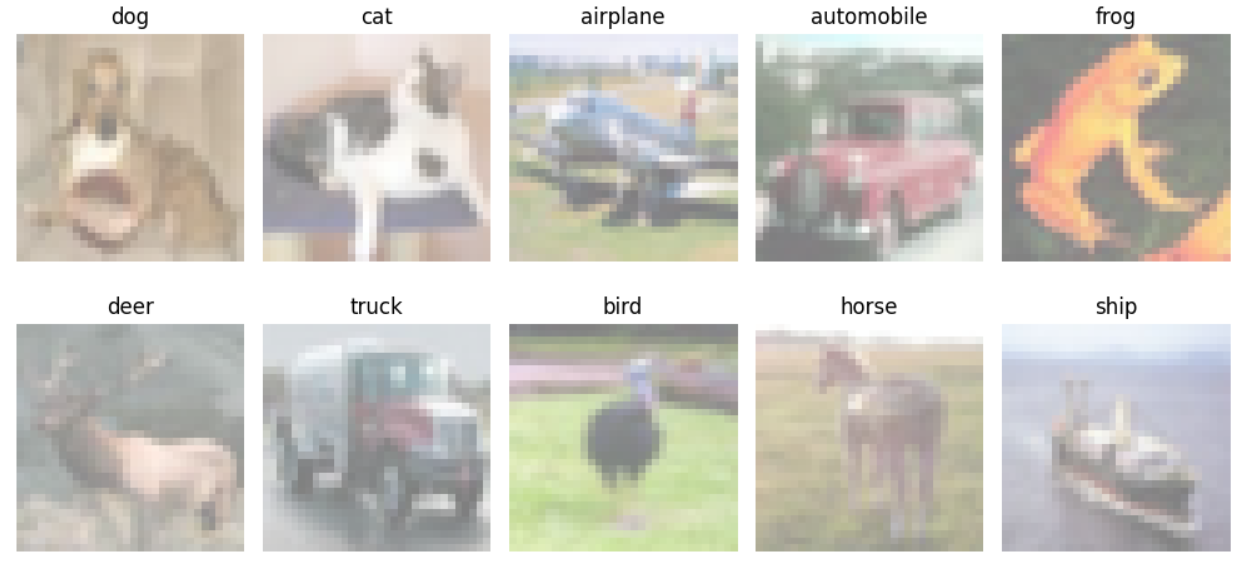}
\caption{Sample images from CIFAR10 dataset}
\label{C1}
\end{figure}

MNIST dataset has been used in the many seminal works of federated learning. It contains images of hand-written digits.
Figure \ref{MNIST Distribution} shows its class distribution, and Figure \ref{Sample MNIST} shows samples from each class. Fashion MNIST dataset has much similarity with MNIST dataset and usually requires more computational time than MNIST to reach a good solution for an algorithm. 
It contains images of 10 types of fashion products.  
The class distribution of this dataset follows a uniform distribution, in particular, all classes contain exactly 6000 images. Figure \ref{Sample FMNIST} shows sample images of the dataset. Compared to the MNIST dataset, Fashion MNIST contains more complex patterns, so nowadays deep learning researchers often use it as a benchmark. MangoLeafBD dataset, a recently released dataset, has not been evaluated in any work of federated learning. It contains 4000 images of mango tree leaves of Bangladesh encompassing seven leaf diseases. Figure \ref{Sample MangoLeadBD} shows sample images different categories of mango leaves. All seven disease categories and the healthy category contain exactly 4000 images. HAM10000 dataset is one of the most popular medical image datasets. The dataset contains 10015 skin cancer images of seven categories. 
Figure \ref{HAM7}, shows sample images from the dataset. The CIFAR-10 dataset has 10 different categories of images of different objects where each category contains 5000 images.
Figure \ref{C1} shows sample images from the dataset. 

In Table \ref{Dataset} we show different characteristics of all five datasets.

\begin{table}[h!]
    \centering
    
    \label{tab:accuracy}
    \begin{tabular}{|c|c|c|c|c|c|c|}
        \hline
        Dataset & Image Type & Size & Data & Train Set & Test Set & Class  \\ 
        \hline
        MNIST \cite{cit29} & Grayscale & 28$\times$28 & 70000 & 60000 & 10000 & 10 \\
        \hline
        Fashion MNIST \cite{cit30} & Grayscale & 28$\times$28 & 70000  & 60000 & 10000 & 10 \\
        \hline
        MangoLeafBD \cite{cit28} & RGB & 240$\times$320 & 4000  & 3200 & 800 & 8 \\
        \hline
        HAM10000 \cite{tschandl2018ham10000} & RGB & 450$\times$600 & 10015  & 8012 & 2003 &  7 \\
        \hline
        CIFAR10 \cite{krizhevsky2009learning} & RGB & 32$\times$32 & 60000 & 48000 & 12000 & 10 \\
        \hline
    \end{tabular}
    \caption{Properties of the datasets}
    \label{Dataset}
\end{table}

\begin{figure}[htbp]
\centering
\includegraphics[scale=0.5]{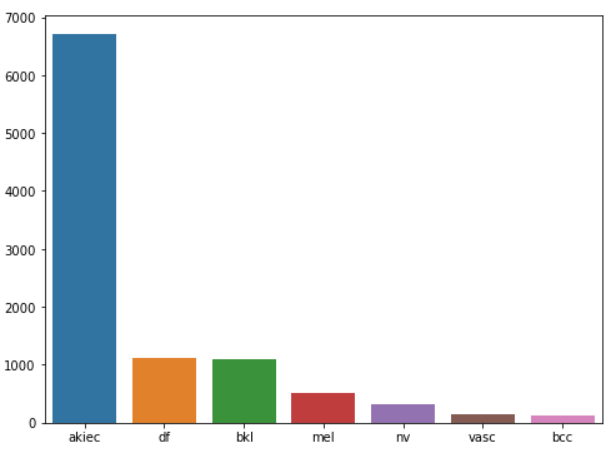}
\caption{Data distribution of HAM10000 dataset}
\label{HAM1}
\end{figure}






\subsection{Experimental Setup}

We now discuss the hardware, the software and the libraries we use to conduct our experiments.
We conduct all the experiments on four different machines as detailed in Table~\ref{HW}. We use both Ubuntu 20.4 and Windows 10 operating systems.  As for the software, all the experiments are implemented in Python 3.9 and conda environment. For deep learning framework, we use Tensorflow\footnote{\url{https://www.tensorflow.org/}}, Keras\footnote{\url{https://keras.io/}} and Pytorch\footnote{\url{https://pytorch.org/}}. The numpy library\footnote{\url{https://numpy.org/}} is  used for different matrix manipulations. Matplotlib\footnote{\url{https://matplotlib.org/}} and Seaborn\footnote{\url{https://seaborn.pydata.org/}} are used to generate the graphs of results. 

\begin{table}[h]
    \centering
    \label{tab:accuracy}
    \begin{tabular}{|c|c|c|c|c|c|c|}
        \hline
        Serial No.  & Processor & SSD & RAM & GPU & Device Type & Operating System  \\ 
        \hline
        1 & Core i7 & 512GB & 32GB & RTx 3080 & Desktop & Ubuntu 22.4\\
        \hline
        2 & Core i5 & 512GB & 16GB  & RTx 3050Ti & Laptop & Windows 10\\
        \hline
        3 & Core i7 & 216GB & 16GB  & None & Desktop  & Windows 10\\
        \hline
        4 & Core i5 & 216GB & 16GB  & None & Laptop  & Windows 10\\
        \hline
    \end{tabular}
    \caption{Different attributes of experimental devices}
    \label{HW}
    
\end{table}

\subsection{Neural Network Models}
We now briefly discuss the neural network models we use in our experiments. 

\subsubsection{Vanilla Neural Network}
\label{sec:vanilla NN MINST FMINST}
For experimenting with MNIST and Fashion MNIST datasets, we, for better comparison with benchmark algorithms, use the same neural network architecture that was used in the original paper of FedAVG algorithm. It is a  four-layer, fully connected neural network where the input layer has 784 neurons and each of the two hidden layers has 200 neurons. The output layer consists of 10 neurons where each of the neurons represents each of the 10 categories of the data. Three activation functions, namely ReLU, are used for transitioning from one layer to the next. This activation function transforms all the negative numbers to 0 and all the non-negative numbers remain the same. No softmax function is used at the end of the output layer.  Categorical cross entropy is used as the loss function, and Adam optimizer is used. The learning rate of the model is set to 0.001.

\subsubsection{RestNet18}
\label{sec:Resnet mangoleafbd}

\begin{figure}[htbp]
\centering
\includegraphics[scale=0.475]{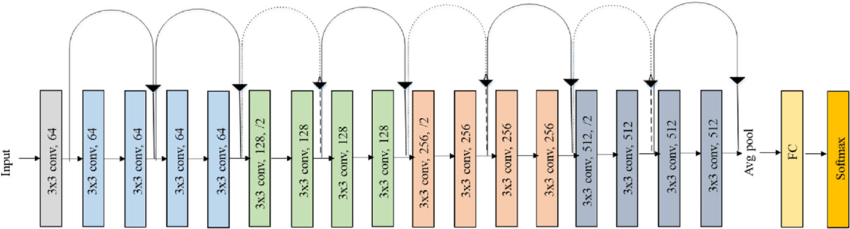}
\caption{ResNet18 architecture \cite{cit27}}
\label{RestNet18}
\end{figure}

For MangoLeafBD dataset, we use a popular image recognition neural network architecture called ResNet18 \cite{cit32} (aka ImageNet). RestNet18 consists of 18 deep convolutional neural networks, and has a residual block which allows it to train the data more effectively. 
Figure \ref{RestNet18} shows the basic diagram of RestNet18. Three convolutional layers have 64 filters each, two convolutional layers with filters of 128, 256, and 512 sizes. Max pooling layers have kernel size $3\times3$ with a stride of 2. There are 28 residual blocks in total. 

\subsubsection{Convolutional Neural Network}
\label{sec:cnn ham100}
For  HAM10000 dataset, we use our own custom-built convolutional neural network whose architecture is described in Table \ref{CNN}.

\begin{table}[h]
    \centering
    
    \label{tab:accuracy}
    \begin{tabular}{|c|c|c|}
        \hline
        Layer/Activation Type  & Number & Kernel/Pool Size  \\ 
        \hline
        Convolutional Layer & 9 & $3\times3$  \\
        \hline
        Max Pooling Layer & 3 & $2\times2$   \\
        \hline
        Dense Layer & 5 & Not Applicable  \\
        \hline
        Flatten Layer & 1 & Not Applicable  \\
        \hline
        ReLU & 15 & Not Applicable  \\
        \hline
        SoftMax & 1 & Not Applicable  \\
        \hline
    \end{tabular}
    \caption{Convolutional neural network description (for HAM10000 dataset)}
    \label{CNN}
\end{table}

\subsubsection{LeNet5}
\label{sec:lenet cifar}
For CIFAR10 dataset, we use a popular network model called LeNet5 \cite{cit37}. It is shown that LeNet-5 yields high-quality result on CIFAR10 in traditional deep learning domain\footnote{\url{https://www.kaggle.com/code/vikasbhadoria/cifar10-high-accuracy-model-build-on-pytorch}}. Table \ref{LN} shows a description of the LeNet-5. 

\begin{table}[h]
    \centering
    
    \label{tab:accuracy}
    \begin{tabular}{|c|c|c|c|c|c|}
        \hline
        Layer Type  & Feature Map &  Size & Kernal Size & Stride & Activation \\ 
        \hline
        Image & 1 & $32\times32$ & - & - & - \\
        \hline
        Convolution & 6 & $28\times28$ & $5\times5$  & 1 & tanh \\
        \hline
        Average Pooling & 6 & $14\times14$ & $2\times2$ & 2 & tanh \\
        \hline
        Convolution & 16 & $10\times10$ & $5\times5$ & 1 & tanh \\
        \hline
        Average Pooling & 16 & $5\times5$ & $2\times2$ & 2 & tanh \\
        \hline
        Convolution &129 & $1\times1$ & $5\times5$ & 1 & tanh \\
        \hline
        FC & - & 84 & - & - & tanh \\
        \hline
        FC & - & 10 & - & - & tanh \\
        \hline
    \end{tabular}
    \caption{LeNet-5 architecture description (for CIFAR-10 dataset)}
    \label{LN}
\end{table}

\subsection{Data Distribution Settings}

In federated learning, data distribution and the number of clients are the two most crucial factors affecting the performance of the algorithms. Therefore, to provide a comprehensive comparison among the existing benchmark algorithms and our proposed models, we experiment with nine combinations of data distribution and client number settings for each dataset which are described next. 

\begin{figure}[htbp]
\centering
\includegraphics[scale=0.7]{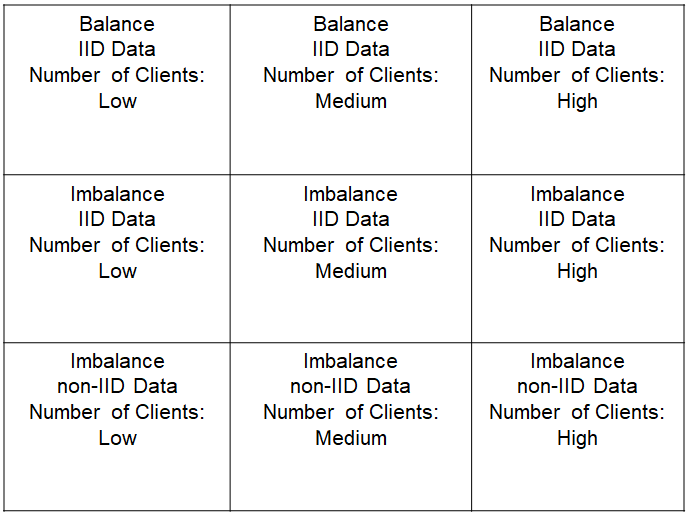}
\caption{Nine settings of experiments}
\label{Temp}
\end{figure}

As discussed earlier, it is known that the ideal case for federated learning is when all clients have the same amount of IID data. So we take this settings and apply the algorithms to different client numbers for observing the performance change pattern. For MNIST, Fashion MNIST, CIFAR10 datasets we use 10, 100, and 600 clients. For HAM10000 dataset we use 4, 10, and 40 clients. For MangoLeafBD dataset we use 8, 32, and 100 clients. Thereafter, we use a setting where clients have imbalanced yet IID data. The number of clients is the same as described above. Finally, we experiment with the most practical, real-life scenario for federated learning where the clients have  imbalanced and non-IID data. Figure \ref{Temp} depicts these nine cases.

\subsection{Results Analysis}
In this section, we analyze the details of results of our experiments on five datasets accross all nine settings. 
Note that both our proposed algorithms yield the same effectiveness (for example, accuracy). Their difference lies in efficiency, i.e., training time. So in the result analysis, we show results of our first algorithm (cf. Section~\ref{sec:proposed algo 1}) and compare it with existing benchmark algorithms. In the efficiency analysis section, we consider both the algorithms.

\subsubsection{MNIST Results}

\begin{figure}[htbp]
\centering
\includegraphics[scale=0.45]{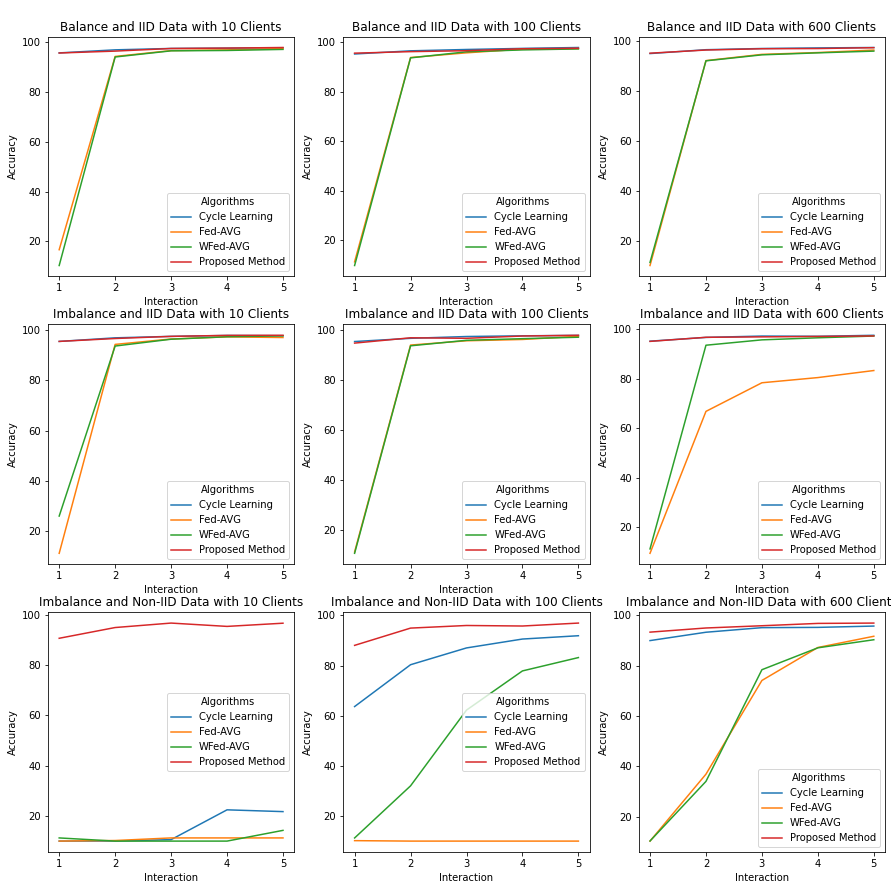}
\caption{Accuracy of MNIST dataset}
\label{MNIST Result}
\end{figure}

As mentioned earlier, we apply a vanilla neural network described in Section~\ref{sec:vanilla NN MINST FMINST} on MINST dataset. This dataset contains 60000 instances in training set, and the numbers of clients we experiment with are 10, 100, and 600. In Figure \ref{MNIST Result}, we show the results across all nine setups. For the case of balance and IID data where each of the clients have the same amount of data and same distribution of classes, after five iterations, all four algorithms reach almost the same level of accuracy irrespective of the number of clients. However, both FedAVG and W-FedAVG take several iterations to reach a good solution while cycle learning and our proposed method reach a good solution after just the first interaction. All the algorithms are able to achieve approximately 99\% accuracy for all algorithms after a certain number of iterations. 

Now let us consider the second row of Figure \ref{MNIST Result} which is imbalanced and IID setting where each client has a different amount of data. 
Here we observe a different scenario.  Cycle learning, W-FedAVG, and our proposed method are still able to achieve approximately 98\% accuracy. For FedAVG, the accuracy starts to decrease when the number of clients increases. FedAVG can achieve 82.65\% accuracy after 5 iterations where the number of the client is 600. But if we see the case of 100 clients, FedAVG achieves 98.25\% accuracy. So the accuracy is dropped by more than 15\% if the client number increased by 500. 

\begin{figure}[htbp]
\centering
\includegraphics[scale=0.45]{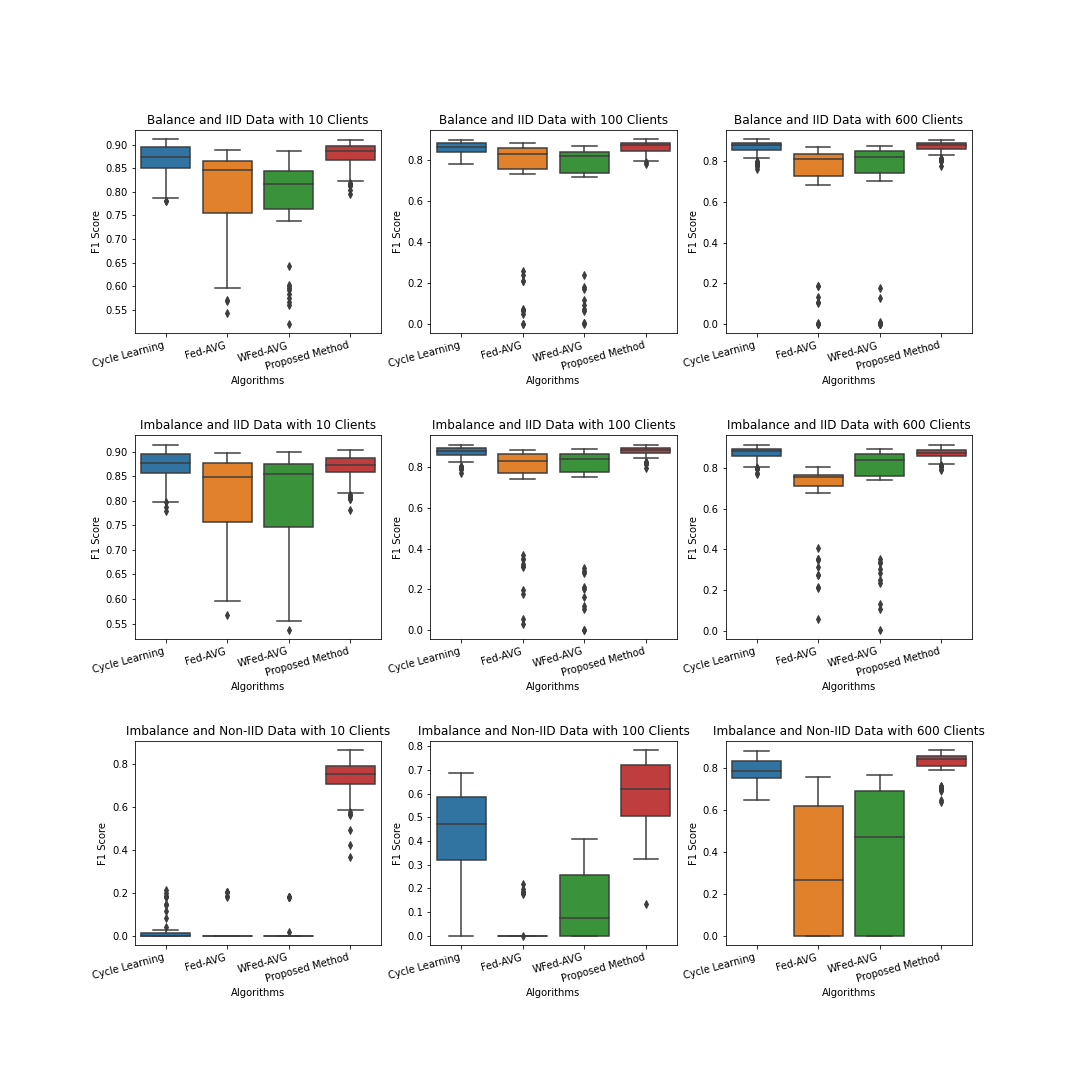}
\caption{F1 score of MNIST dataset}
\label{m1f}
\end{figure}

Now let us draw our attention to the last row of Figure \ref{MNIST Result}. In this setting, the clients have different amount of data and clients have only one type of class data. 
For example, when the client number is 10, each client has data of a single digit only. In this scenario, we see that our proposed model outperforms the other three algorithms. Although cycle learning is able to reach 23.13\% accuracy after 5 iterations, our algorithm achieves 90\% accuracy from the very first iteration. In the end, our algorithm is able to reach 99\% accuracy. In this setting, if we start increasing the number of clients, then the data sensitivity issue starts to manifest in existing benchmark algorithms. If we increase the client number from 10 to 100, then cycle learning and W-FedAVG start to improve their results. In the 5th iteration, cycle learning and W-FedAVG achieve more than 90\% and 80\% accuracy respectively, whereas FedAVG does not improve. When we increase the client number from 100 to 600, we can see that FedAVG also starts to improve, as well as cycle learning and W-FedAVG.  To conclude, in MNIST dataset, we can claim that our proposed method outperforms the existing benchmark algorithms as it is found to be performing well in all the scenarios. Figure~\ref{m1f} demonstrates F1 scores of MNIST dataset.

\subsubsection{Fashion MNIST Results}

\begin{figure}[htbp]
\centering
\includegraphics[scale=0.45]{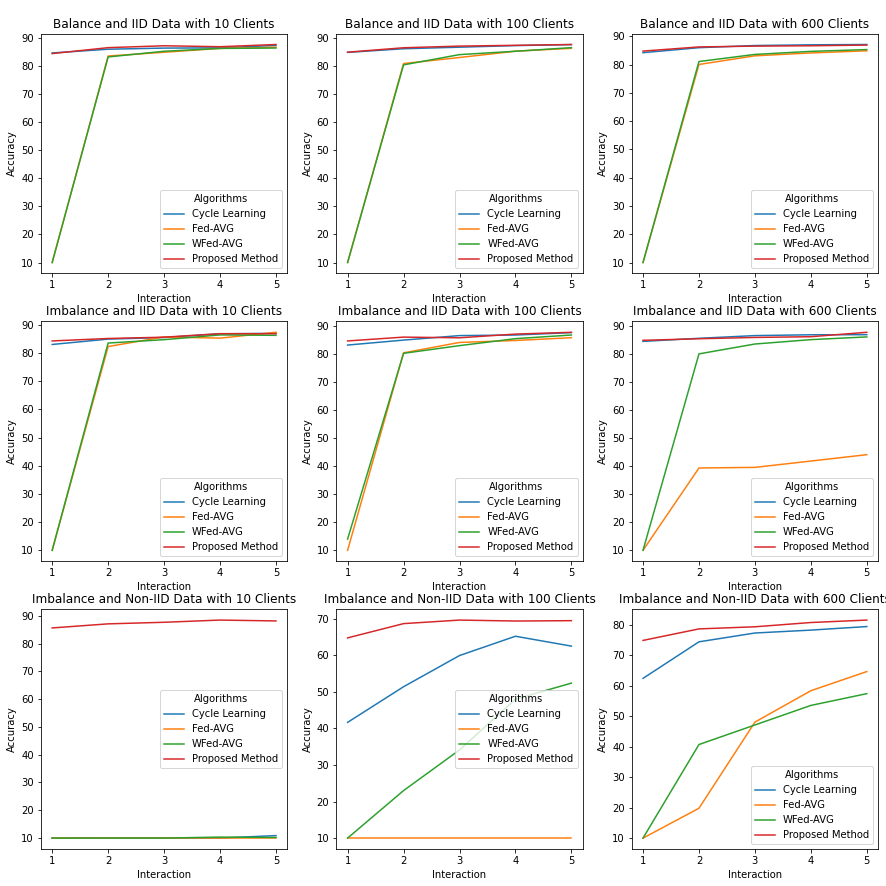}
\caption{Accuracy of Fashion MNIST dataset}
\label{FMNIST Result}
\end{figure}

Fashion MNIST dataset contains 60000 instances in training set. Here we use the same neural network model that we used for MNIST dataset, and the numbers of clients we experiment with are 10, 100, and 600. As Fashion MNIST dataset possesses a more complex pattern as compared to the MNIST dataset, the accuracy of all the algorithms is comparatively lower. In Figure \ref{FMNIST Result}, for the first row, accuracy is above 80\% in all three cases. However, FedAVG and W-FedAVG need multiple iterations to achieve a satisfactory accuracy while our proposed model and cycle learning achieve good accuracy from the first iteration. So using our proposed method or cycle learning can  reach an acceptable solution relatively quickly. 

The case is different if we focus on the three graphs of the second row in Figure \ref{FMNIST Result}. In this case, the clients have different amounts of data. 
We notice that if we increase the number of clients from 10 to 100, accuracy of FedAVG and W-FedAVG decreases slightly, although after some more iterations, both are able to pass 80\% accuracy. But when we increase clients such as 600, FedAVG fails to go above 50\% accuracy, whereas others achieve 80\%. 

\begin{figure}[htbp]
\centering
\includegraphics[scale=0.45]{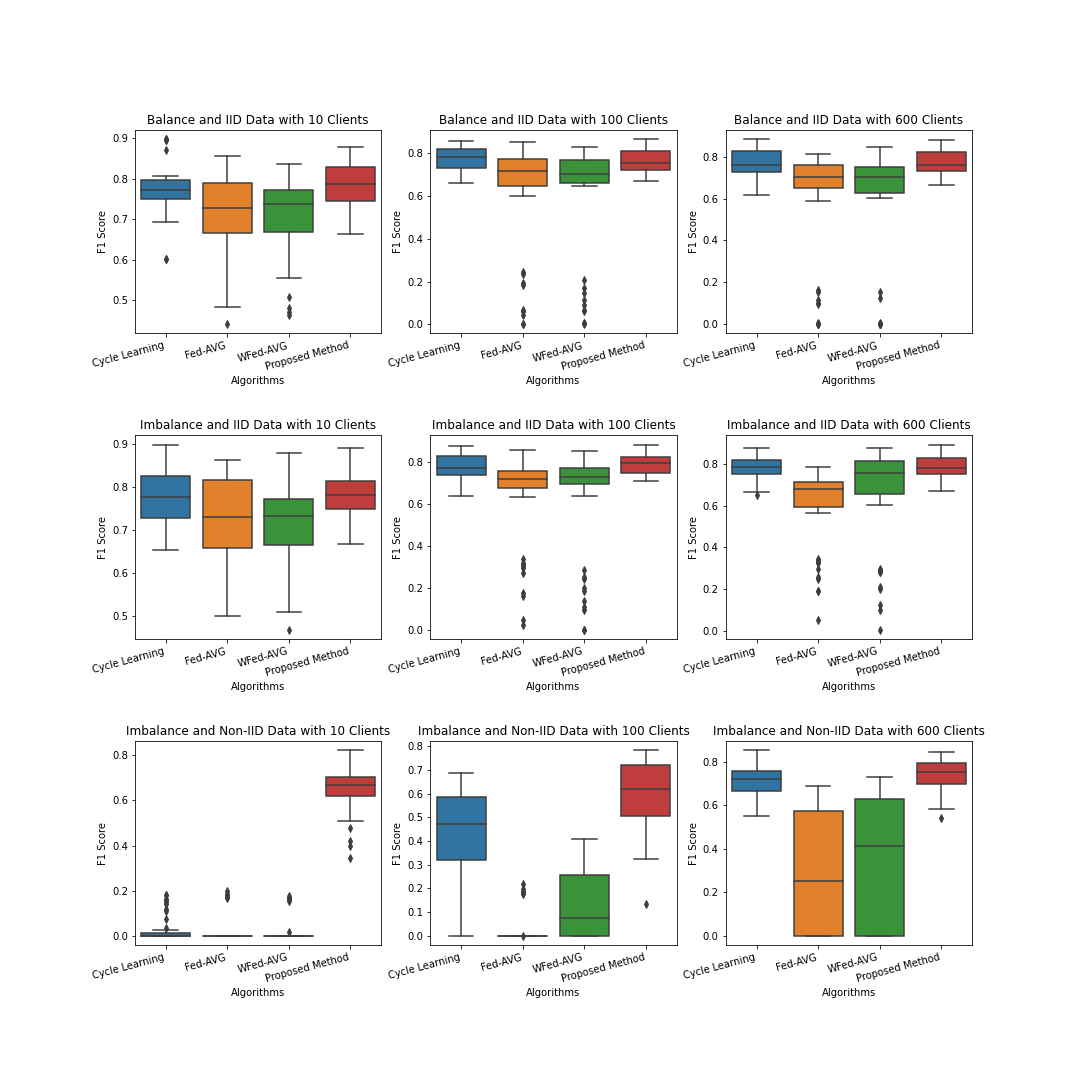}
\caption{F1 score of Fashion MNIST dataset}
\label{m2f}
\end{figure}

Now if we focus on the last row of Figure \ref{FMNIST Result}, we see that only our proposed method gives satisfactory results, and the other three demonstrate a mediocre performance. However, if we start increasing the number of clients, W-FedAVG and cycle learning both start to achieve better results. FedAVG also starts improving when the client number is increased to 600. But in this case also, only our proposed method achieves more than 80\% accuracy in 5 iterations. The closest one to us is cycle learning which is able to achieve 78.67\% accuracy. So in this case, also we can claim that our proposed method shows better results than the existing benchmark algorithms. Figure~\ref{m2f} demonstrates F1 scores of Fashion MNIST dataset.

\subsubsection{MangoLeafBD Results}

\begin{figure}[htbp]
\centering
\includegraphics[scale=0.45]{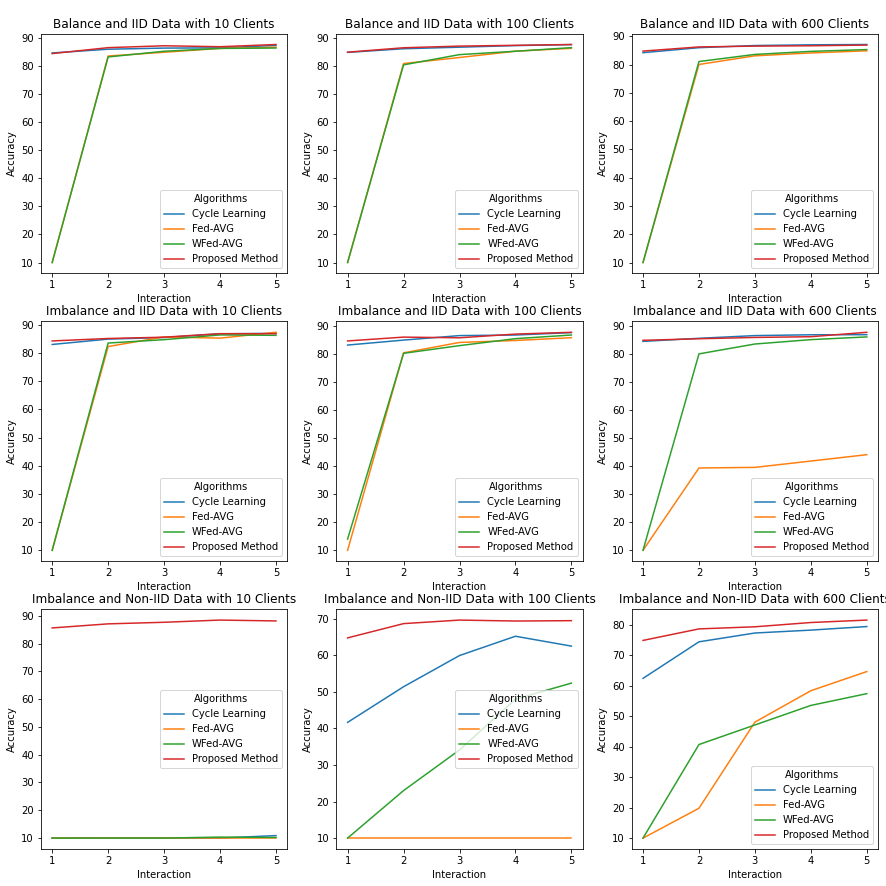}
\caption{Accuracy of MangoleafBD dataset}
\label{Mango Result}
\end{figure}

For MangoLeafBD dataset, we use RestNet18 (cf. section~\ref{sec:Resnet mangoleafbd}) as our model.  There are 3200 images in the training set, and the client numbers we experiment with are 8, 32, and 100.
Compared to other datasets, the images of  MangoLeafBD are larger. We see in Figure \ref{Mango Result} that like MNIST and Fashion MNIST datasets, here cycle learning and our proposed method reach 95\% accuracy in very early iterations. FedAVG and W-FedAVG also achieve this level of accuracy, but they take more time. 

In the second row of Figure \ref{Mango Result} where imbalanced yet IID data scenario holds, by increasing the number of clients from 32 to 100, FedAVG performance significantly drops. While other algorithms behave roughly similarly as the first row. 

\begin{figure}[htbp]
\centering
\includegraphics[scale=0.45]{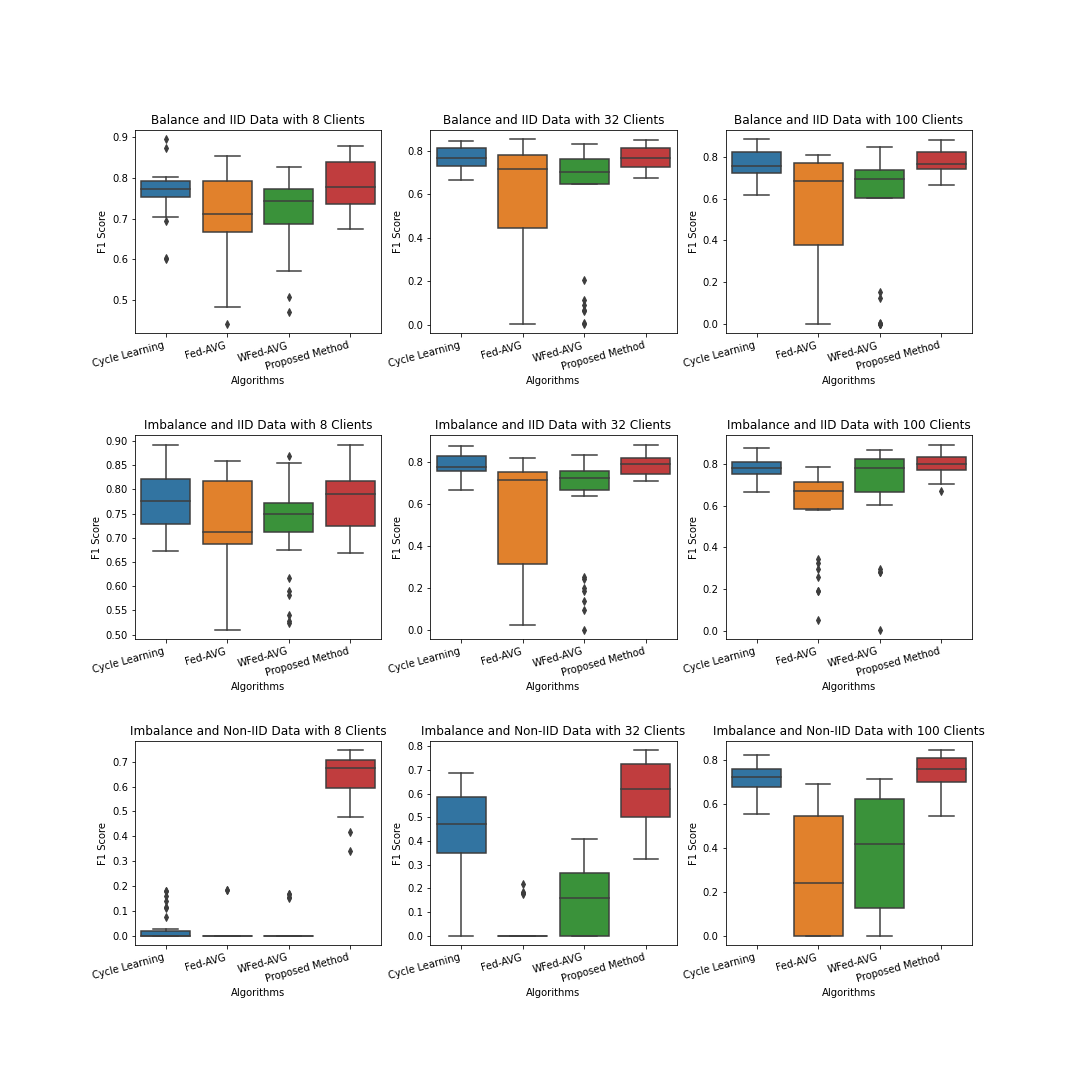}
\caption{F1 score of MangoLeafBD dataset}
\label{m3f}
\end{figure}

In the last line of Figure \ref{Mango Result} where imbalanced and non-IID data scenario holds, only our proposed method gives satisfactory results, and the other three yield poor performance. But if we start increasing the number of clients, W-FedAVG and cycle learning both demonstrate an increase in their performance, but this is still considered to be mediocre. Only our proposed method is able to achieve more than 90\% accuracy in all the non-IID settings for MangoLeafBD dataset. 
Figure~\ref{m3f} demonstrates F1 scores of MangoLeafBD dataset.

\subsubsection{HAM10000 Results}

\begin{figure}[htbp]
\centering
\includegraphics[scale=0.45]{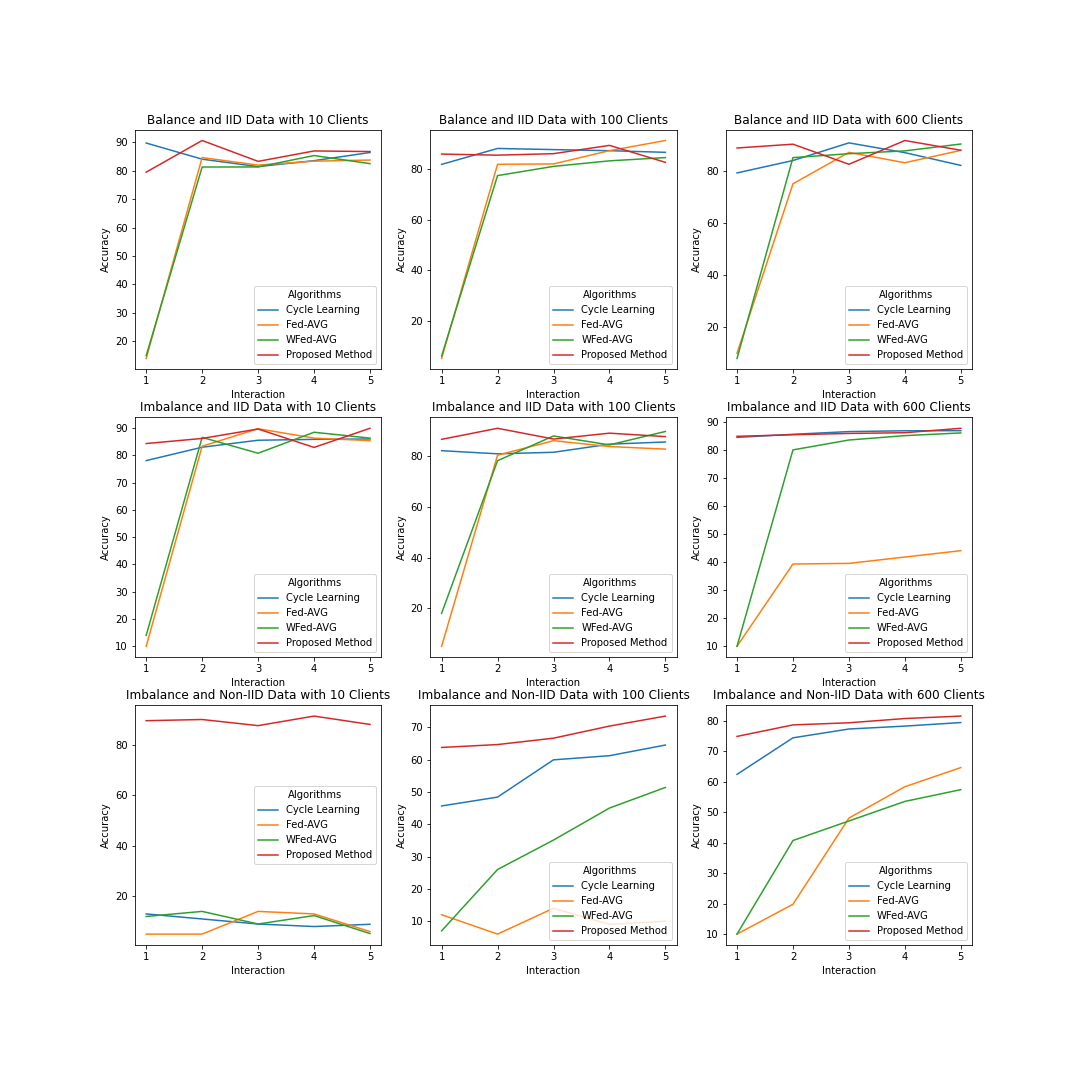}
\caption{Accuracy of HAM10000 dataset}
\label{HAcc}
\end{figure}

\begin{figure}[htbp]
\centering
\includegraphics[scale=0.45]{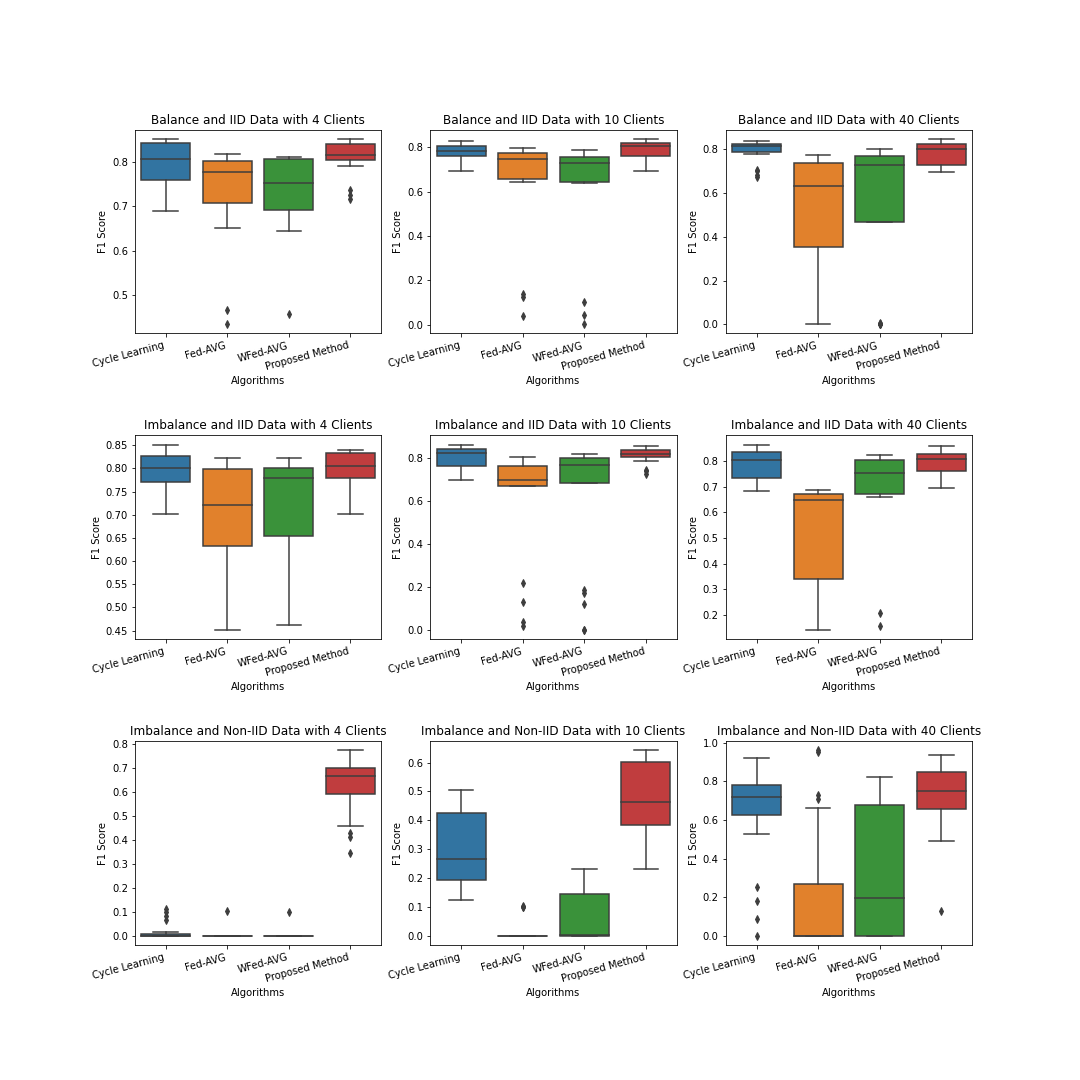}
\caption{F1 score of HAM10000 dataset}
\label{m4f}
\end{figure}

For HAM10000 dataset we use the CNN model described in Section~\ref{sec:cnn ham100}. There are 8012 data instances in the training set of HAM10000 dataset, and the number of clients we experiment with are 4, 10, and 40. In Figure \ref{HAcc}, we observe that 
when the data is balanced and IID data, then almost every algorithm receives satisfactory result. But when the data is imbalanced and Non-IID only our methodology gives a satisfactory result. 
Figure~\ref{m4f} shows the F1 scores of HAM10000 dataset.

\subsubsection{CIFAR10 Results}

\begin{figure}[htbp]
\centering
\includegraphics[scale=0.45]{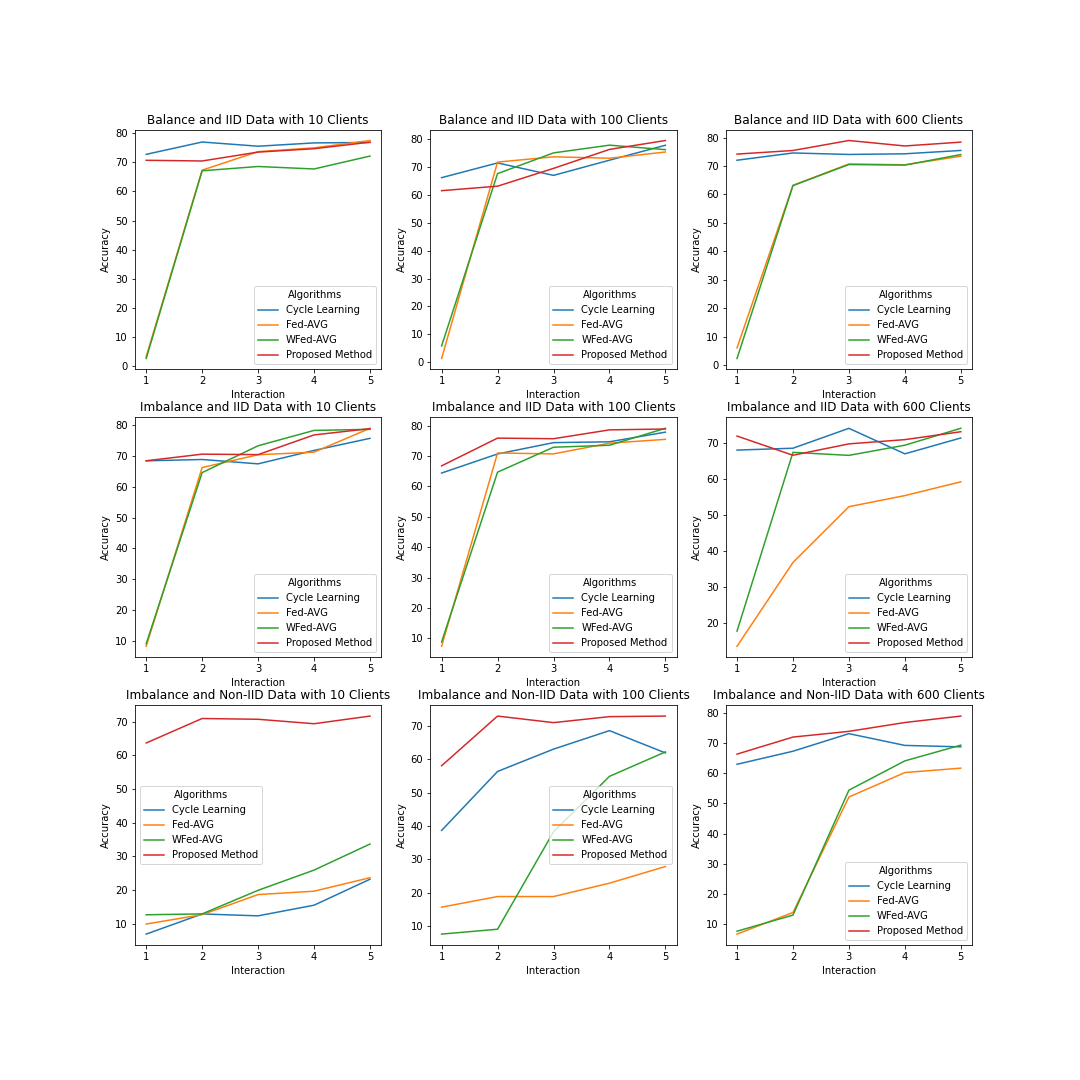}
\caption{Accuracy of CIFAR10 dataset}
\label{CAcc}
\end{figure}

For the CIFAR10, we use LeNet-5 (cf. Section~\ref{sec:lenet cifar}) as our learning model. Here the number of instances in training set is 48000, and the number of clients we experiment with are 10, 100, and 600. In Figure \ref{CAcc}, we see that like the other four datasets, cycle learning and our proposed method reach high accuracy after the very first iteration. FedAVG and W-FedAVG also achieve this, but only after taking a certain time. In the second row of Figure \ref{CAcc} where the data are imbalanced yet IID, by increasing the client number from 100 to 600, performance of FedAVG deteriorates, and the other algorithms' behaviour remains largely similar  as the first row. In the last row of Figure \ref{CAcc} where the data are imbalanced and non-IID, only our proposed method demonstrates good performance. However, after starting to  increase the number of clients, W-FedAVG and cycle learning both witness gradually  increasing performance, albeit not at the level of that of ours. Only our proposed method is able to achieve more than 70\% accuracy in all the non-IID settings for CIFAR-10 dataset. 


\subsection{Efficiency Analysis}

We now take into account the time that our algorithms require to complete the training process, and compare it with that of existing benchmark algorithms.

As mentioned earlier, the key difference between our proposed two algorithms lies in efficiency, i.e., training time. Our method 1 (cf. Section~\ref{sec:proposed algo 1}) is comparatively more efficient than method 2 (cf. Section~\ref{sec:proposed algo 2}) as the former employs parallel training. For our efficiency analysis, we conduct our experiments in one single computer with one dataset and four clients. The data distribution is randomized. For method 1, we experiment with the hyper-parameter parallel window sizes of 1, 2, and 4.

\begin{table}[h]
    \centering
    \label{tab:accuracy}
    \begin{tabular}{|c|c|c|c|c|c|c|c|}
        \hline
          Datasets & M1(P=1) &  M1(P=2) & M1(P=4) & M2 & FedAVG & W-FedAVG & CL \\ 
        \hline
        MNIST & 4.44 & 3.12 & 2.19 & 5.19 & 1.54 & 2.43 & 4.07 \\
        \hline
        Fashion-MNIST & 8.54 & 6.02 & 4.41 & 9.01 & 3.31 & 3.43 & 8.14 \\
        \hline
        MangoLeafBD & 15.23 & 9.30 & 6.18 & 14.58 & 5.44 & 6.00 & 14.38 \\
        \hline
        HAM10000 & 18.14 & 13.00 & 9.36 & 19.28 & 9.56 & 9.49 & 17.22 \\
        \hline
        CIFAR10 & 11.35 & 8.54 & 6.27 & 11.11 & 5.59 & 6.09 & 10.55 \\
        \hline
    \end{tabular}
    \caption{Training time of proposed algorithms and benchmark algorithms. M1: proposed method 1 and M2: proposed method 2. $P$ indicates the parallel window size). CL stands for cycle learning.}
    \label{TAT}
\end{table}

\begin{figure}[htbp]
\centering
\includegraphics[scale=0.65]{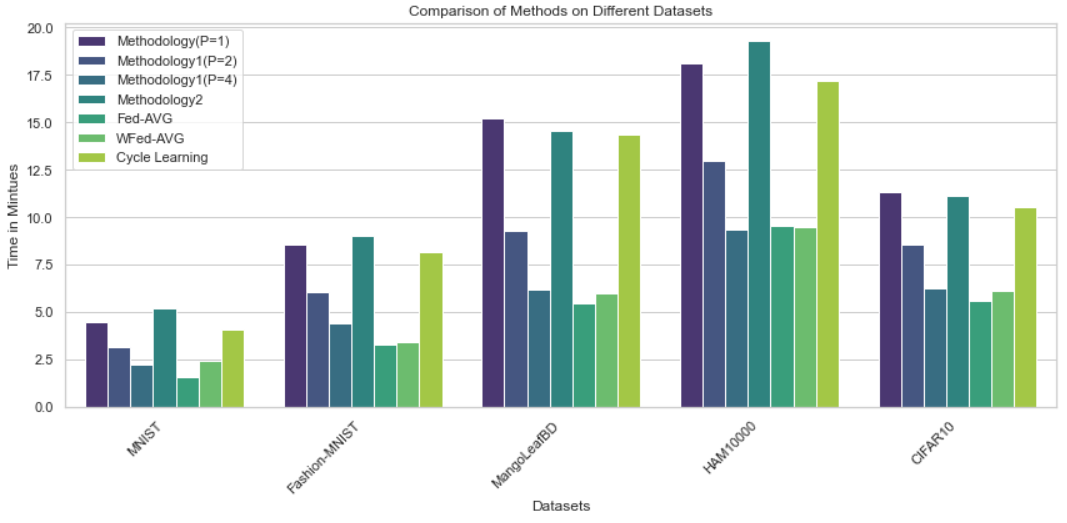}
\caption{Training time of proposed algorithms and benchmark algorithms. $P$ indicates the parallel window size.}
\label{TAF}
\end{figure}

In Table \ref{TAT}, our proposed methods 1 and 2 are denoted by M1 and M2 respectively, and CL stands for cycle learning. The value of $P$ indicates the parallel window size of method 1. In Table \ref{TAT} and Figure \ref{TAF}, we see that training time of proposed method 1 decreases when we increase our parallel window size $P$, which is expected. When it comes to efficiency, FedAVG outperforms all other algorithms. The reason for FedAVG's superiority is that FedAVG allows parallel training at its full potential. When we fix our parallel window size $P$ to 4, which is the maximum for this simulation environment, we achieve almost the same efficiency as FedAVG. On the other hand, our proposed method 2 takes the highest time in the training process because the forward propagation is performed sequentially by the clients, i.e., multiple clients cannot train at the same time in this method. In this method, in order to reduce the computational load from the server, we have to compromise the training time.



\subsection{Discussion}


Let us focus on the Figures \ref{MNIST Result}, \ref{FMNIST Result}, \ref{Mango Result},  \ref{HAcc}, and \ref{CAcc}. We see that accuracy is different for different datasets. That said, the algorithms' performance pattern seems to look largely similar in all five figures. When all clients have the same amount of data and data are IID (i.e., balanced and IID setting -- the first row of the figures), all the algorithms perform relatively well. Among them, FedAVG and W-FedAVG need multiple iterations to achieve good performance. The main reason for this, we argue, is: FedAVG and W-FedAVG take the average of the weights and biases of the local models. As the initial weights and biases are assigned randomly, different local models react differently in back-propagation as their data are different. Thus the global model gets a mediocre solution. After multiple iterations, however, the global model gradually overcomes this limitation. Now the question arises: why cycle learning and our proposed model are found to be performing well in very early iterations? The answer, we conjecture, is that in cycle learning algorithm one model keeps getting updated while traveling to all the clients. Therefore, each client improves the results of their previous clients. So after finishing the iteration, the model gains knowledge from all the  clients. As for our proposed algorithm, we have already explained in detail in Section~\ref{sec:our proposed algo motivation} that it starts incorporating patterns of the data of all the clients from the very beginning, thanks to its idea of micro-level parallel training inspired by traditional mini-batch technique of neural networks.

Now let us consider the second setting, i.e., imbalanced and IID data, (second row) of Figures \ref{MNIST Result}, \ref{FMNIST Result}, \ref{Mango Result},  \ref{HAcc}, and \ref{CAcc}. The performance of FedAVG gradually decreases as we increase the number of clients. The main reason behind this, we argue, is that here the clients have different amounts of data. So the local models do not have the same importance, but still, FedAVG considers all the local models the same while aggregating the models' weights and biases. 
W-FedAVG algorithms tackles this problem by performing weighted average instead of plain average. The weight of a client is proportional to the amount of data that the client has. For the case of cycle learning, this is not an issue because the model visits every client sequentially. In our proposed method, the server, in the first place, knows the clients' amount of data and accordingly instructs the clients to divide their data into roughly equal sized groups. The clients train their local models in parallel on these equal sized groups. That is why  proposed method does not face this issue.

Now let us consider the imbalanced and non-IID setting (third row) of Figures \ref{MNIST Result}, \ref{FMNIST Result}, \ref{Mango Result},  \ref{HAcc}, and \ref{CAcc}. In all these figures we see that except our proposed method, other algorithms are unable to achieve good accuracy when the number of clients is small. In this case, FedAVG and W-FedAVG yield poor performance, because the clients' data are non-IID, i.e., a client may not have data of all class labels. Thus the local models have a high bias towards the particular class labels it contain. 
If if we consider our proposed method and how it tackles this problem, then the answer lies in the same arguments placed in Section~\ref{sec:our proposed algo motivation}, i.e., it its novel parallelism process that incorporates the idea of traditional mini-batch algorithm. In non-IID data setting, a client may have data of only a few class labels. But in the proposed method we take the loss from multiple clients at a time. It is not only giving the advantage of training time, but also gives the data patterns from different class labels. That is why in the process of back-propagation, our method is able to ensure that the loss is not biased towards a particular class label of the dataset. In this way, our proposed method achieves good results where other federated learning algorithm are unable to do so. 


Our proposed methods have the following limitations. The proposed methods have to rely on clients' loss values and parameters. So a faulty client can harm the global model greatly. Client dependency needs to be safer. Also, In our experiments we assume the communication error to be zero. In real life, however, there can be communication delays or problems while transferring a model or loss. However, this area is out of scope of our current research.

From our theoretical understanding and experimental result, it is safe to claim that our proposed algorithm outperforms the existing benchmark algorithms in the case of imbalanced and non-IID data settings. 

\section{Conclusion}
\label{sec:conclusion}
Due to the concern about data privacy, large machine learning systems often cannot be built as they require data from multiple sources. The purpose of introducing federated learning is to build systems where machine learning models can be trained in users' end devices without their data being shared, thereby securing the data privacy of the users. However, the existing federated learning systems yield relatively lower accuracy as compared to traditional learning, and works well only in some specific situation such as in IID data scenario. The purpose of this research has been to propose a solution that helps neural network-based federated learning systems to be more accurate regardless of the data distribution settings. Experimental results of our proposed algorithms on a number of popular datasets suggest that our proposed algorithm improves the accuracy over the benchmark algorithms. 





This research also opens the door for exploring several other aspects of federated learning systems. Firstly, our semi-centralized federated learning system takes a larger time for training despite having the advantage of reduced server dependency. Semi-centralized federated learning system with faster training is a domain for further investigation. Secondly, we have assumed that all clients are technically faultless. However, faculty client identification is important for real-life deployment of federated learning systems. Thirdly, we do not take into account of communication delays in our research which is an important topic for future research.

%
%
%
%
\bibliographystyle{plain}
\bibliography{paper}
\end{document}